\documentclass[11pt]{article}
\usepackage{ersr_paper}

\usepackage{amsmath,amsfonts,bm}

\def\eqref#1{equation~\ref{#1}}

\def\1{\bm{1}}

\DeclareMathAlphabet{\mathsfit}{\encodingdefault}{\sfdefault}{m}{sl}
\SetMathAlphabet{\mathsfit}{bold}{\encodingdefault}{\sfdefault}{bx}{n}

\usepackage{graphicx}
\usepackage[most]{tcolorbox}
\usepackage{adjustbox}
\usepackage{booktabs}
\usepackage{multirow}
\usepackage{tabularx}

\title{Expected Reasoning-Step Return Unifies On-Policy Learning from Rewards and Teachers}

\author{%
  \textbf{Qiangqiang He}\textsuperscript{1} \quad
  \textbf{Jin Li}\textsuperscript{2}\\[0.20em]
  \small \textsuperscript{1}State Key Laboratory for Novel Software Technology, Nanjing University, Nanjing, China\\[-0.10em]
  \small \textsuperscript{2}College of Software Engineering, Southeast University, Nanjing, China\\[0.20em]
  \small \texttt{qqh@smail.nju.edu.cn} \quad \texttt{jin\_li@seu.edu.cn}%
}
\date{}
\begin{document}

\maketitle

\begin{abstract}
On-policy reasoning models can learn from task rewards or teacher signals, but these sources differ in form and can favor conflicting updates, leaving unclear which should guide a given reasoning action. We introduce \textbf{Expected Reasoning-Step Return (ERSR)}, which treats semantic reasoning steps as macro-actions and uses Monte Carlo student-policy rollouts to estimate the expected final task reward of student-generated and teacher-proposed actions in a common return space for step-level comparison. ERSR analysis reveals an outcome-dependent asymmetry: student actions are more beneficial than teacher replacements on successful trajectories, whereas teacher replacements become more beneficial on failed trajectories. We further show that student answer-probe gains track student-step ERSR utility and distinguish beneficial from harmful reasoning steps. Based on these findings, we propose \textbf{Return-Referenced On-Policy Learning (R$^2$OPL)}, which reinforces student reasoning on successful trajectories and distills teacher signals on failed ones, while using group success rate for difficulty scaling and student-probe gains for step-level modulation. Experiments across reasoning benchmarks and teacher--student configurations show that R$^2$OPL consistently outperforms strong baselines. ERSR training dynamics further show that R$^2$OPL jointly exploits substantial utility from both reward- and teacher-side signals, whereas existing hybrids often leave substantial residual utility in one branch.

\end{abstract}

\section{Introduction}

Recent advances in LLM reasoning increasingly rely on on-policy post-training, where models learn from trajectories sampled by their current policy. Reward-driven methods such as GRPO\citep{shao2024deepseekmath}, DAPO\citep{yu2026dapo}, and GSPO\citep{zheng2025groupsequencepolicyoptimization} reinforce behaviors associated with successful solutions, while on-policy distillation (OPD) provides dense teacher supervision on student-visited states\citep{agarwal2024policy, li2026rethinking, ko2026scaling}. Yet these signals need not align: teacher supervision may undervalue correct student reasoning, favor teacher-preferred over student-beneficial actions, or become unreliable under teacher--student mismatch and noisy supervision\citep{akhondzadeh2026reward, yang2026beyond, li2026rethinking, ko2026scaling, ding2026does}. This raises a fundamental question: when reward- and teacher-derived signals disagree, which should guide the optimization of a reasoning action?

\begin{figure}[t]
    \centering
    \includegraphics[width=\linewidth]{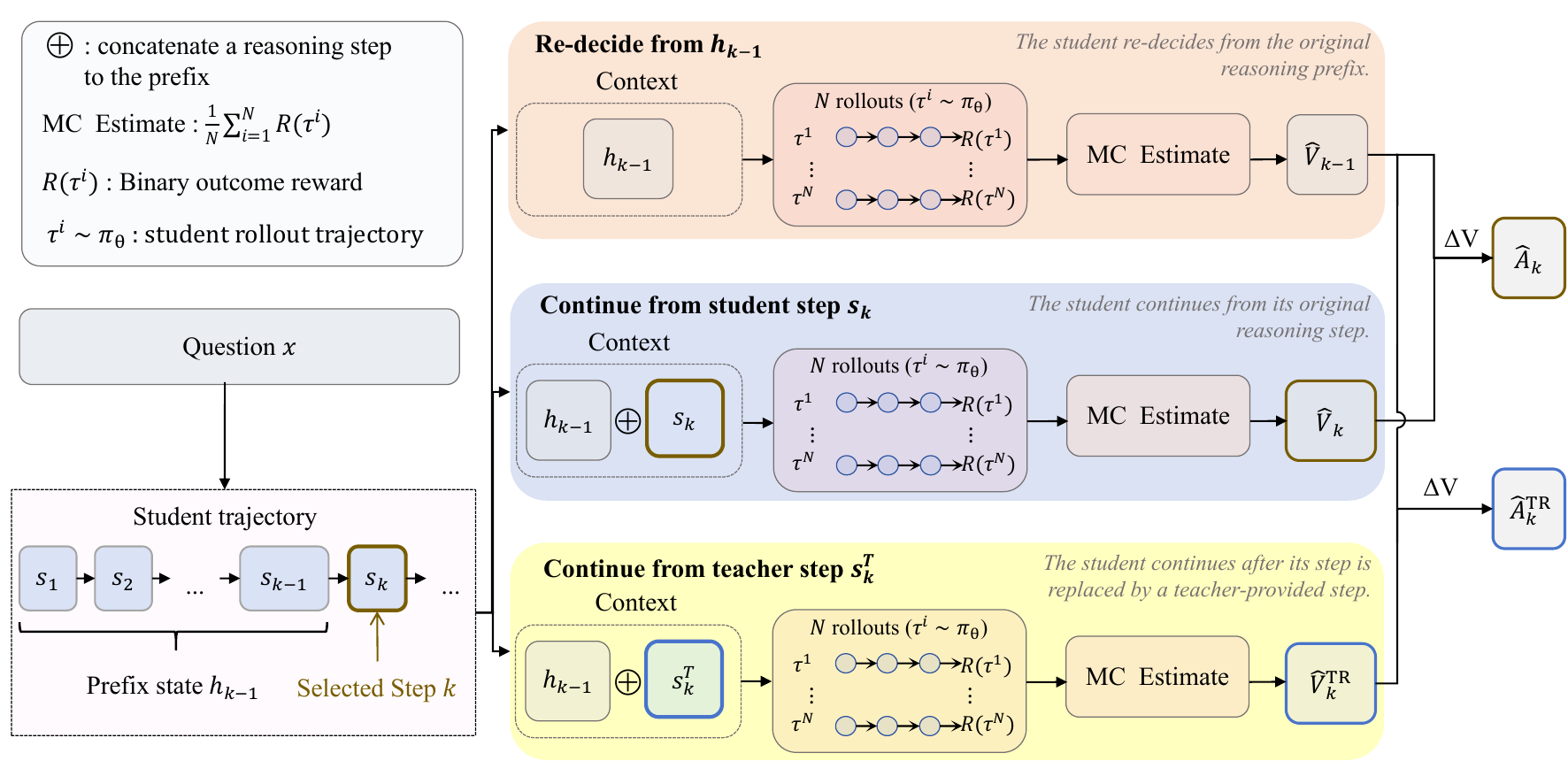}
\caption{
ERSR evaluates a reasoning step by estimating its student-step advantage $\widehat{A}_k$ and teacher-replacement advantage $\widehat{A}_k^{\mathrm{TR}}$, with all subsequent continuations sampled from the student policy.
}
    \label{fig:ersr}
\end{figure}

Despite their different objectives, reward-based reinforcement and teacher-guided distillation ultimately seek to improve the student's expected task reward. Reward-based methods optimize this objective directly, whereas OPD uses teacher--student discrepancy as a surrogate for beneficial learning. Yet disagreement is only a coarse indicator of learning utility\citep{wang2026not}: even stronger teachers can provide limited or negative transfer when their signals are incompatible with the student\citep{wu2026lightning, lu2026strong}, and teacher likelihood may diverge from task-level correctness\citep{zhang2026beyond}. Thus, what matters is not how strongly the teacher disagrees with the student, but whether the favored reasoning action increases the student's expected task reward.

To directly measure this utility, we introduce \textbf{Expected Reasoning-Step Return (ERSR)}, which treats each semantic reasoning step as a macro-action and evaluates its contribution to the expected final task reward. As illustrated in Figure~\ref{fig:ersr}, from the same prefix state, we consider three continuations: the student re-decides from the original prefix, continues from its original reasoning step, or continues after replacing that step with a teacher-provided step. All subsequent rollouts are sampled from the student policy, and Monte Carlo estimation yields the expected final reward $\widehat{V}$ of each continuation. Their differences define the \emph{student-step advantage} $\widehat{A}_k$ and \emph{teacher-replacement advantage} $\widehat{A}_k^{\mathrm{TR}}$, placing student and teacher actions in the same expected-return space.

Offline ERSR analysis reveals a clear outcome-dependent asymmetry: on successful trajectories, student reasoning steps are more beneficial than teacher replacements, whereas on failed trajectories, teacher replacements become substantially more beneficial than the student's original actions. We further find that student answer-probe gains consistently track student-step ERSR utility across both outcomes, while signals computed on observed student steps do not reliably reveal teacher-replacement utility. Motivated by these findings, we propose \textbf{Return-Referenced On-Policy Learning (R$^2$OPL)}, which reinforces student reasoning on successful trajectories and distills teacher signals on failed ones, while using group success rate for problem-level difficulty scaling and student-probe gains for fine-grained reasoning-step modulation.

Experiments across eight benchmarks spanning mathematical, scientific, and logical reasoning show that R$^2$OPL consistently outperforms strong reward-based, distillation, and hybrid baselines. ERSR training dynamics show that R$^2$OPL jointly exploits both reward- and teacher-side utility, whereas hybrid methods often leave substantial residual utility in one branch. Our contributions are threefold: (1) we introduce ERSR, which compares and tracks student-step and teacher-replacement utility in a common expected-return space; (2) using ERSR, we reveal an outcome-dependent asymmetry between the two and show that student answer probes track student-step utility; and (3) we translate these findings into R$^2$OPL, which unifies on-policy learning from rewards and teacher signals through outcome-conditioned routing, difficulty scaling, and step-level modulation.

\section{Understanding Reward and Teacher Signals with ERSR}

\subsection{Preliminaries}

\paragraph{On-policy learning.}
We use \emph{on-policy learning (OPL)} as a broad term for methods that learn from trajectories sampled from the current student policy, including both reward-based reinforcement learning and on-policy distillation.
Let $x \sim \mathcal{D}$ denote a reasoning problem and
$\tau=(y_1,\ldots,y_T)\sim\pi_\theta(\cdot\mid x)$ a trajectory sampled from the student policy $\pi_\theta$.
Each trajectory receives a binary outcome reward $R(\tau)\in\{0,1\}$ indicating whether its final answer is correct.
The ultimate objective of OPL is to maximize the student's expected task reward,
\begin{equation}
    J_{\mathrm{OPL}}(\theta)
    =
    \mathbb{E}_{x\sim\mathcal{D},\,\tau\sim\pi_\theta(\cdot\mid x)}
    \left[R(\tau)\right].
    \label{eq:expected_reward}
\end{equation}
Reward-based reinforcement learning and on-policy distillation both seek to improve $J_{\mathrm{OPL}}$ in Eq.~\ref{eq:expected_reward}, but construct their learning signals differently.
For a unified view, we express their token-level learning signals through a common policy-gradient-style surrogate.
Let $h_t=(x,y_{<t})$ denote the current context, $\widehat{A}_t$ the learning signal assigned to token $y_t$, and $\operatorname{sg}[\cdot]$ the stop-gradient operator:
\begin{equation}
    J_{\mathrm{PG}}(\theta;\widehat{A})
    =
    \mathbb{E}_{x\sim\mathcal{D},\,\tau\sim\pi_\theta(\cdot\mid x)}
    \left[
        \sum_{t=1}^{T}
        \operatorname{sg}\!\left[\widehat{A}_t\right]
        \log \pi_\theta(y_t\mid h_t)
    \right].
    \label{eq:pg_objective}
\end{equation}

\paragraph{Reinforcement learning.}
Reward-based RL constructs learning signals directly from task outcomes. A representative example is GRPO\citep{shao2024deepseekmath}, which samples a group of trajectories $\{\tau_i\}_{i=1}^{G}$ for the same prompt and normalizes their outcome rewards to obtain
\begin{equation}
    \widehat{A}^{\mathrm{GRPO}}_i
    =
    \frac{R_i-\overline{R}}
    {\sigma_R+\epsilon},
    \qquad
    \overline{R}
    =
    \frac{1}{G}\sum_{j=1}^{G}R_j,
    \qquad
    \sigma_R
    =
    \sqrt{
        \frac{1}{G}
        \sum_{j=1}^{G}
        \left(R_j-\overline{R}\right)^2
    }.
    \label{eq:rl_advantage}
\end{equation}
The resulting trajectory-level advantage is shared across all tokens in $\tau_i$. While this provides a direct signal from task rewards, its credit assignment remains inherently coarse, as different reasoning steps may contribute differently to the final reward.

\paragraph{On-policy distillation.}
OPD learns from a teacher policy $\pi_T$ on states visited by $\pi_\theta$. A standard setting in OPD assumes that the teacher achieves a higher expected task reward than the student,
\begin{equation}
    \mathbb{E}_{x\sim\mathcal{D},\,\tau\sim\pi_T(\cdot\mid x)}
    [R(\tau)]
    >
    \mathbb{E}_{x\sim\mathcal{D},\,\tau\sim\pi_\theta(\cdot\mid x)}
    [R(\tau)].
    \label{eq:teacher_superiority}
\end{equation}
Under this premise, OPD uses the teacher--student log-probability difference as the surrogate learning signal in Eq.~\ref{eq:pg_objective} for dense token-level supervision on student-visited states,
\begin{equation}
    \widehat{A}^{\mathrm{OPD}}_t
    =
    \log \pi_T(y_t\mid h_t)
    -
    \log \pi_\theta(y_t\mid h_t).
    \label{eq:opd_advantage}
\end{equation}
This signal assigns larger values to tokens favored more by the teacher than by the student. However, the teacher's higher expected reward does not guarantee that its local preferences improve the student's expected task reward; teacher--student disagreement may instead reflect preferences unrelated to task utility\citep{yang2026beyond, ding2026does}. Consequently, $\widehat{A}^{\mathrm{OPD}}_t$ remains a surrogate learning signal, and a larger distillation signal need not imply greater expected-return utility.

\subsection{Expected Reasoning-Step Return}

\paragraph{Reasoning steps as actions.}
To compare reward- and teacher-driven learning in a common task space, we evaluate their favored reasoning actions by their effect on the student's expected final reward. Token-level returns are formally well defined, but individual tokens rarely represent complete reasoning operations, teacher--student differences may reflect lexical variation, and token-level effects are often too small for efficient Monte Carlo (MC) estimation. A detailed analysis of action granularity for ERSR is provided in Appendix~\ref{app:token_return}. We therefore treat each semantic reasoning step as a macro-action, enabling direct comparison between the student's original step and a teacher-provided alternative in the same expected-return space. Steps are identified from decoded responses using structural and semantic boundaries and mapped back to the original token sequence while preserving the original tokenization, with details provided in Appendix~\ref{app:step_segmentation}.

\paragraph{ERSR definition and estimation.}
Consider a student trajectory segmented into reasoning steps $\tau=(s_1,\ldots,s_K)$, and let $h_{k-1}=(x,s_{<k})$ denote the prefix before step $k$. As illustrated in Figure~\ref{fig:ersr}, we compare three continuations from this common prefix: the student re-decides from $h_{k-1}$, continues from its original step $s_k$, or continues after replacing $s_k$ with a teacher-provided step $s_k^T$. Denoting $h_k=(h_{k-1},s_k)$ and $h_k^T=(h_{k-1},s_k^T)$, their true expected final rewards are\begin{equation}
\begin{aligned}
    V_{k-1} &= V^{\pi_\theta}(h_{k-1}), &
    V_k &= V^{\pi_\theta}(h_k), &
    V_k^T &= V^{\pi_\theta}(h_k^T), \\
    A_k &= V_k-V_{k-1}, &
    A_k^{\mathrm{TR}} &= V_k^T-V_{k-1}.
\end{aligned}
\label{eq:ersr_definition}
\end{equation}
Here, all subsequent continuations are sampled from the same student policy $\pi_\theta$. The \emph{student-step advantage} $A_k$ measures the expected benefit of preserving the student's reasoning step relative to re-decision from the same prefix, while the \emph{teacher-replacement advantage} $A_k^{\mathrm{TR}}$ measures that of replacing it with the teacher-provided step. Together, they provide a task-grounded basis for comparing student reinforcement and teacher distillation. Since these values are not directly observable, we estimate them using $N$ independent student-policy continuations,
\begin{equation}
    \widehat{V}_N(h)
    =
    \frac{1}{N}\sum_{n=1}^{N}R\!\left(\tau^{(n)}\right),
    \qquad
    \tau^{(n)}\sim\pi_\theta(\cdot\mid h).
    \label{eq:ersr_mc}
\end{equation}
Substituting the corresponding MC estimates into Eq.~\ref{eq:ersr_definition} yields $\widehat{A}_k$ and $\widehat{A}_k^{\mathrm{TR}}$. We use MC@$N$ to denote estimation with $N$ student-policy rollouts. ERSR thus places student and teacher reasoning actions in a common expected-return space.

\subsection{Empirical Findings with ERSR}

We analyze ERSR using Qwen3-1.7B as the student and Qwen3-4B-Instruct-2507\citep{yang2025qwen3} as the teacher on the mathematical dataset DAPO-17K\citep{yu2026dapo}. We evaluate 30,000 reasoning steps and use MC@128 to estimate the student-step advantage $\widehat{A}$ and teacher-replacement advantage $\widehat{A}^{\mathrm{TR}}$. Additional details on data sampling and evaluation are provided in Appendix~\ref{app:ersr_evaluation_pipeline}.

\begin{table*}[t]
\centering
\small
\setlength{\tabcolsep}{7.0pt}
\renewcommand{\arraystretch}{1.12}

\caption{
Outcome-dependent ERSR utility.
Positive, zero, and negative denote the percentages of steps by utility sign; $\Delta_{\mathrm{sign}}$ is the positive--negative percentage-point gap.
}
\label{tab:ersr_outcome_utility}

\begin{tabular}{@{}lcccccc@{}}
\toprule
\textbf{Outcome}
& \textbf{ERSR Utility}
& \textbf{Mean}
& \textbf{Positive (\%)}
& \textbf{Zero (\%)}
& \textbf{Negative (\%)}
& $\boldsymbol{\Delta_{\mathrm{sign}}}$ \textbf{(\%)} \\
\midrule

\multirow{2}{*}{$R=1$}
& $\widehat{A}$
& \textbf{$+0.032$}
& \textbf{51.6}
& 15.9
& 32.6
& \textbf{$+19.0$} \\

& $\widehat{A}^{\mathrm{TR}}$
& $+0.007$
& 44.8
& 14.0
& 41.2
& $+3.7$ \\

\midrule

\multirow{2}{*}{$R=0$}
& $\widehat{A}$
& $-0.021$
& 30.2
& 26.8
& 43.0
& $-12.9$ \\

& $\widehat{A}^{\mathrm{TR}}$
& \textbf{$+0.064$}
& \textbf{49.8}
& 24.2
& 26.0
& \textbf{$+23.7$} \\

\bottomrule
\end{tabular}
\end{table*}

\paragraph{Outcome-dependent action utility.}
Table~\ref{tab:ersr_outcome_utility} reveals a clear outcome-dependent asymmetry in step utility across both mean advantage and sign distribution. On successful trajectories, student actions yield a higher mean advantage than teacher replacements, $+0.032$ versus $+0.007$, together with a stronger positive sign balance of $+19.0\%$ versus $+3.7\%$, indicating greater utility for preserving student reasoning. On failed trajectories, the preference reverses: student actions have a negative mean advantage of $-0.021$, whereas teacher replacements reach $+0.064$, with sign balances of $-12.9\%$ and $+23.7\%$, respectively. These results support outcome-conditioned source selection: successful trajectories favor student reinforcement, while failed trajectories benefit more from teacher-guided correction. Further analyses across teachers and problem difficulty are provided in Appendix~\ref{app:ersr_teacher_difficulty}.

\paragraph{Answer probes track student-step utility.}
Trajectory outcome helps select the learning source, but does not distinguish the utility of individual reasoning steps. Since estimating ERSR at every step requires expensive MC rollouts, we seek a lightweight proxy whose variation tracks step-level ERSR utility. For each reasoning state $h_k$, we use the fixed answer template $c_{\mathrm{probe}}$: \verb|Therefore, the answer is \boxed{<answer>}.|, with \verb|<answer>| replaced by the reference answer $a^\ast$. If $a^\ast$ tokenizes as $(z_1,\ldots,z_M)$, we define

\begin{equation}
    P_k^{\mathrm{probe}}
    =
    \frac{1}{M}\sum_{m=1}^{M}
    \pi_{\mathrm{probe}}(z_m\mid h_k,c_{\mathrm{probe}},z_{<m}),
    \qquad
    \Delta P_k=P_k^{\mathrm{probe}}-P_{k-1}^{\mathrm{probe}},
    \label{eq:answer_probe_gain}
\end{equation}
where $\pi_{\mathrm{probe}}$ denotes the probing model. We then evaluate whether $\Delta P_k$ tracks the corresponding MC-estimated ERSR utility.

\begin{table*}[t]
\centering
\scriptsize
\setlength{\tabcolsep}{1.8pt}
\renewcommand{\arraystretch}{1.10}

\caption{
\textbf{Pearson correlations} between step-level signals and ERSR utility. $\Delta P^{T-S}=\Delta P^T-\Delta P^S$ and $P^{T-S}=P^T-P^S$.
The highest absolute correlation per utility group is \textbf{bolded}; shaded cells mark negative teacher-side correlations on successful trajectories.
}
\label{tab:probe_ersr_correlation}

\begin{tabular*}{\textwidth}{@{\extracolsep{\fill}}ccccccccccc@{}}
\toprule
&
\multicolumn{4}{c}{$\boldsymbol{\widehat{A}}$}
&
\multicolumn{6}{c}{$\boldsymbol{\widehat{A}^{\mathrm{TR}}}$}
\\
\cmidrule(lr){2-5}
\cmidrule(lr){6-11}

\textbf{Outcome}
& $\Delta P^S$
& $\Delta P^T$
& $e^{D_{\mathrm{TS}}}$
& $e^{L_T}$
& $\Delta P^S$
& $\Delta P^T$
& $\Delta P^{T-S}$
& $P^{T-S}$
& $e^{D_{\mathrm{TS}}}$
& $e^{L_T}$
\\
\midrule

$R=1$
& \textbf{0.246}
& 0.178
& \cellcolor{gray!15}$-0.092$
& \cellcolor{gray!15}$-0.104$
& 0.003
& 0.007
& 0.005
& 0.038
& $\mathbf{-0.071}$
& $-0.048$
\\

$R=0$
& \textbf{0.211}
& 0.131
& $+0.109$
& $+0.089$
& 0.008
& $-0.013$
& $\mathbf{-0.022}$
& 0.016
& $-0.012$
& $-0.002$
\\

\bottomrule
\end{tabular*}

\end{table*}

\begin{figure*}[t]
    \centering
    \begin{tabular}{@{}c@{\hspace{6pt}}c@{}}
        \includegraphics[width=0.48\textwidth]{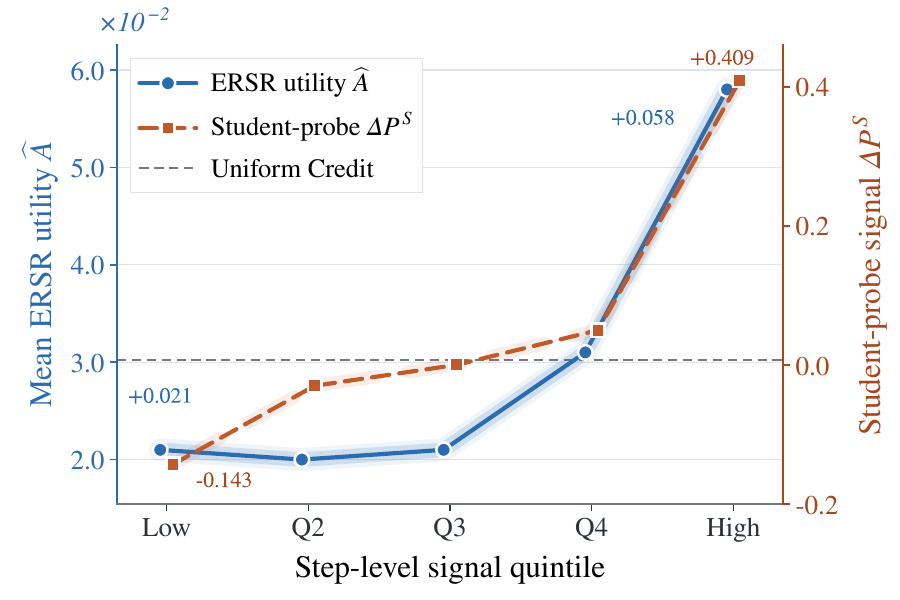}
        &
        \includegraphics[width=0.48\textwidth]{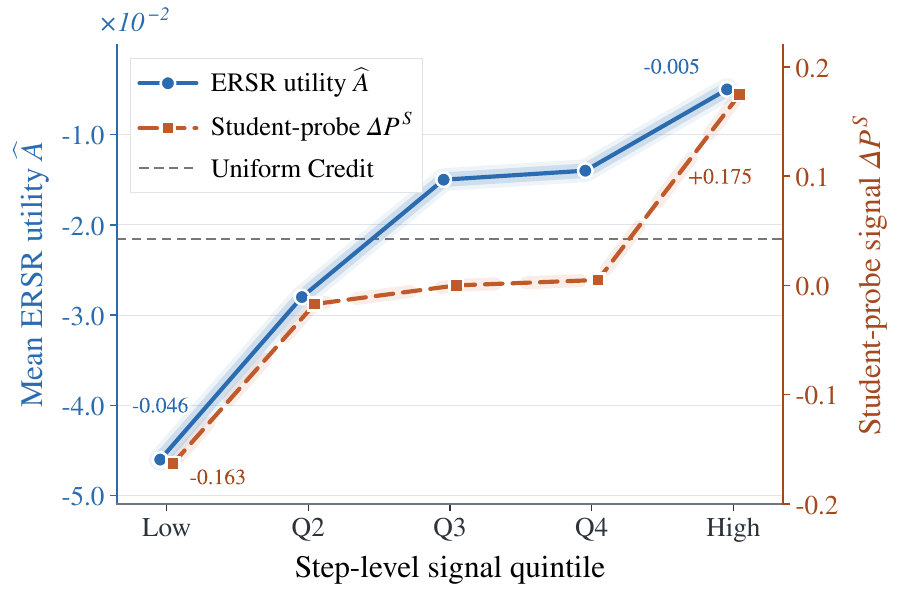}
        \\[-1pt]
    {\small (a) Student-probe ranking on successful trajectories}
    &
    {\small (b) Student-probe ranking on failed trajectories}
    \end{tabular}
    \caption{
        \textbf{Student-probe ranking of step utility.}
        (a) Student-step utility across probe quintiles on successful trajectories.
        (b) Student-step utility across probe quintiles on failed trajectories.
    }
    \label{fig:probe_credit}
\end{figure*}

We instantiate $\pi_{\mathrm{probe}}$ with either the student policy $\pi_\theta$ or the teacher policy $\pi_T$, yielding student- and teacher-probe gains $\Delta P_k^S$ and $\Delta P_k^T$, both evaluated on the same student-generated reasoning steps. For comparison with conventional teacher-side signals, we define the step-mean teacher log probability as $L_k^T=\frac{1}{|s_k|}\sum_{t\in s_k}\log\pi_T(y_t\mid h_t)$ and the step-mean teacher--student discrepancy as $D_k^{\mathrm{TS}}=\frac{1}{|s_k|}\sum_{t\in s_k}\left[\log\pi_T(y_t\mid h_t)-\log\pi_\theta(y_t\mid h_t)\right]$, and use their probability-space forms $e^{L_k^T}$ and $e^{D_k^{\mathrm{TS}}}$ in the correlation analysis. Table~\ref{tab:probe_ersr_correlation}, together with Figure~\ref{fig:probe_credit}, reveals three findings:

\begin{itemize}
    \item \textbf{Student probes best track student-step utility.}
    The student-probe gain $\Delta P^S$ exhibits the strongest correlation with $\widehat{A}$ under both outcomes, reaching $0.246$ on successful trajectories and $0.211$ on failed trajectories, compared with $0.178$ and $0.131$ for the teacher probe. Figure~\ref{fig:probe_credit} further shows that $\Delta P^S$ distinguishes more beneficial steps on successful trajectories and more harmful steps on failed trajectories, supporting its use for fine-grained modulation within the outcome-selected learning branch.

    \item \textbf{ERSR exposes a mismatch between teacher preference and task utility.}
    Among teacher-side evaluations of the student action, $\Delta P^T$ correlates most strongly with $\widehat{A}$, while $e^{D_{\mathrm{TS}}}$ and $e^{L_T}$ are substantially weaker. Notably, both become negatively correlated with $\widehat{A}$ on successful trajectories, at $-0.092$ and $-0.104$, respectively, despite being weakly positive on failed trajectories. From the ERSR perspective, this suggests that teacher--student discrepancy and teacher likelihood may primarily reflect teacher-specific preferences rather than reward-aligned reasoning utility, consistent with prior observations that teacher preference need not align with reward\citep{yang2026beyond,ding2026does}.

    \item \textbf{Observed student-step signals do not reveal teacher-replacement utility.}
    In contrast to $\widehat{A}$, none of the examined probe, likelihood, or discrepancy signals shows meaningful correlation with $\widehat{A}^{\mathrm{TR}}$, with all absolute correlations below $0.08$. Thus, signals computed from the observed student step do not reliably recover the utility of a teacher replacement.
\end{itemize}

\section{Method}

\subsection{Return-Referenced On-Policy Learning}

Building on the ERSR findings, we propose \textbf{Return-Referenced On-Policy Learning (R$^2$OPL)}, a lightweight on-policy method that unifies learning from task rewards and teacher signals within a single objective while retaining their distinct roles. For each prompt $x\sim\mathcal{D}$, we sample a group of $G$ responses $\{\tau_i\}_{i=1}^{G}$ from the student policy $\pi_\theta$, where $\tau_i=(y_{i,1},\ldots,y_{i,T_i})$ and each response receives a binary reward $R_i\in\{0,1\}$. Let $d_x$ denote the prompt-level difficulty weight and $\widehat{A}^{\mathrm{R^2OPL}}_{i,t}$ the branch-specific learning signal for token $y_{i,t}$, both defined below. Following the policy-gradient formulation in Eq.~\ref{eq:pg_objective}, R$^2$OPL maximizes
\begin{equation}
    J_{\mathrm{R^2OPL}}(\theta)
    =
    \mathbb{E}_{\substack{
        x\sim\mathcal{D},\,
        \{\tau_i\}_{i=1}^{G}\sim\pi_\theta(\cdot\mid x)
    }}
    \left[
        \frac{d_x}{G}
        \sum_{i=1}^{G}
        \frac{1}{T_i}
        \sum_{t=1}^{T_i}
        \operatorname{sg}\!\left[
            \widehat{A}^{\mathrm{R^2OPL}}_{i,t}
        \right]
        \log\pi_\theta(y_{i,t}\mid h_{i,t})
    \right].
    \label{eq:r2opl_objective}
\end{equation}

\paragraph{Outcome-conditioned routing and difficulty scaling.}
Following the outcome-dependent utility revealed by ERSR, R$^2$OPL routes successful trajectories to reward-based reinforcement and failed trajectories to teacher-guided OPD. We further use the group success rate $\overline{R}_x$ to adapt the overall update strength according to prompt difficulty:
\begin{equation}
    \widehat{A}^{\mathrm{R^2OPL}}_{i,t}
    =
    \begin{cases}
        \mu\,\widehat{A}^{\mathrm{RL}}_{i,t}, & R_i=1,\\
        \lambda\,\widehat{A}^{\mathrm{OPD}}_{i,t}, & R_i=0,
    \end{cases}
    \qquad
    \overline{R}_x=\frac{1}{G}\sum_{i=1}^{G}R_i,
    \qquad
    d_x=1-\overline{R}_x.
    \label{eq:r2opl_routing_scaling}
\end{equation}
Here, $\mu$ and $\lambda$ control the relative optimization scales of the reinforcement and distillation branches. The difficulty weight $d_x$ assigns stronger updates to prompts the current student rarely solves and progressively attenuates both branches as the group success rate increases.

\paragraph{Step-level modulation with student probes.}
R$^2$OPL is optimized at the token level, whereas answer-probe gains are defined over semantic reasoning steps. We segment each response $\tau_i$ into steps $(s_{i,1},\ldots,s_{i,K_i})$ using the ERSR procedure and let $k(t)$ denote the step containing token $t$, so all tokens within a step share the same modulation factor. Following the ERSR analysis, we use the student-probe gain $\Delta P^S_{i,k}$ to characterize the utility of the student's own reasoning step.

On successful trajectories, Figure~\ref{fig:probe_credit}(a) shows that student-step utility remains positive across probe quintiles and rises with $\Delta P^S$, so larger probe gains identify more beneficial steps. We therefore use the binary reward as the RL advantage and reinforce steps with larger probe gains:
\begin{equation}
    \widehat{A}^{\mathrm{RL}}_{i,t}
    =
    R_i\left(
        1+\alpha_R\Delta P^{S}_{i,k(t)}
    \right),
    \qquad R_i=1,
    \label{eq:r2opl_rl_advantage}
\end{equation}
where $\alpha_R$ controls the modulation strength. The binary reward provides the base credit, while the probe modulates its magnitude according to student-step utility.

On failed trajectories, Figure~\ref{fig:probe_credit}(b) shows that student-step utility remains negative across probe quintiles but becomes progressively less negative as $\Delta P^S$ increases, indicating that lower probe gains identify more detrimental reasoning steps. We therefore retain the token-level OPD signal and assign stronger teacher supervision to steps with lower probe gains:
\begin{equation}
    \widehat{A}^{\mathrm{OPD}}_{i,t}
    =
    \left[
        \log\pi_T(y_{i,t}\mid h_{i,t})
        -
        \log\pi_\theta(y_{i,t}\mid h_{i,t})
    \right]
    \left(
        1-\alpha_D\Delta P^{S}_{i,k(t)}
    \right),
    \qquad R_i=0,
    \label{eq:r2opl_opd_advantage}
\end{equation}
where $\alpha_D$ controls the modulation strength. Lower student-probe gains induce stronger distillation for more harmful steps, while higher probe gains yield weaker teacher correction.

\begin{figure}[t]
    \centering
    \begin{tabular}{@{}c@{\hspace{4pt}}c@{}}
        \includegraphics[height=0.26\textwidth,keepaspectratio]{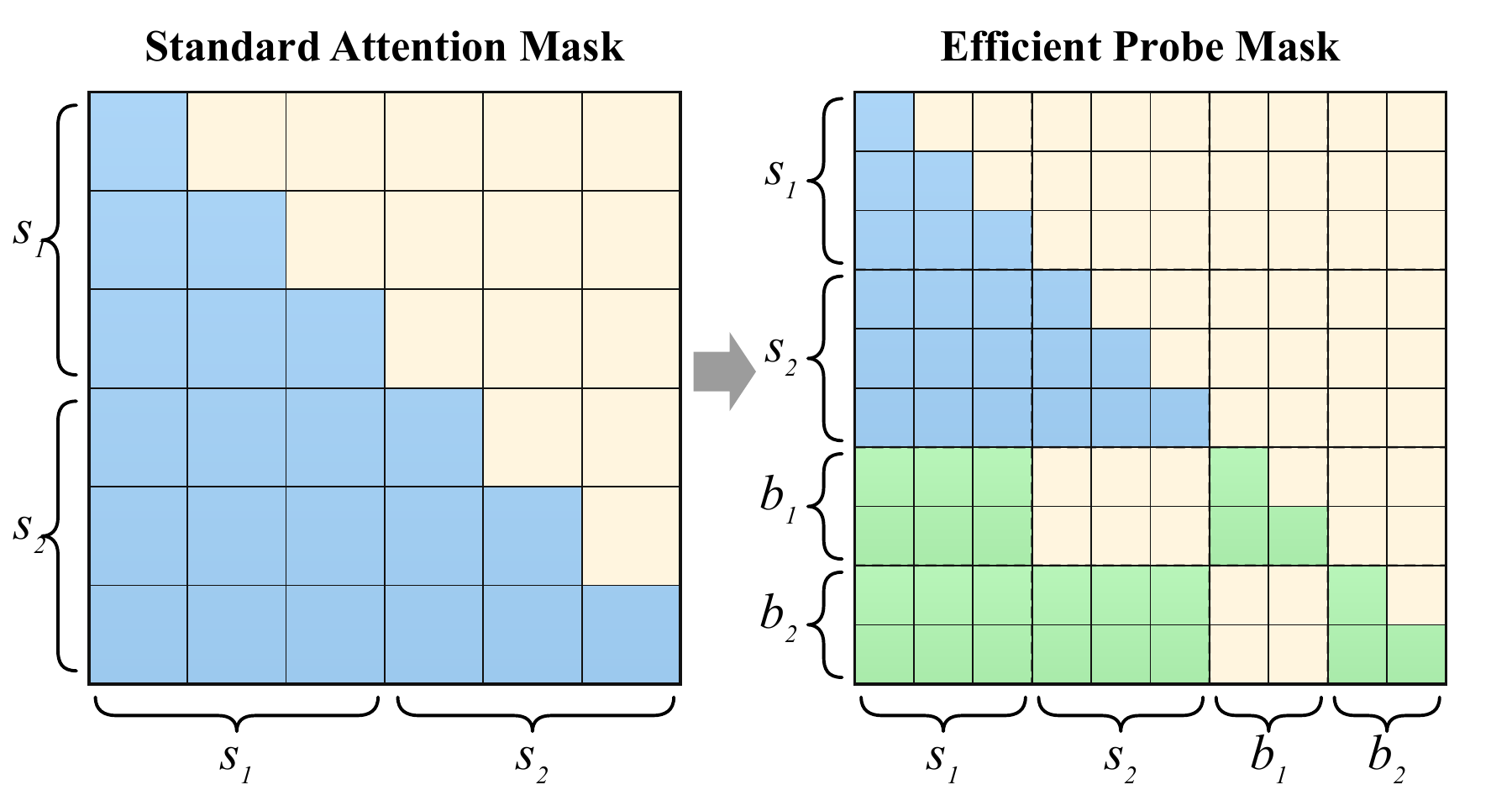}
        &
        \includegraphics[height=0.26\textwidth,keepaspectratio]{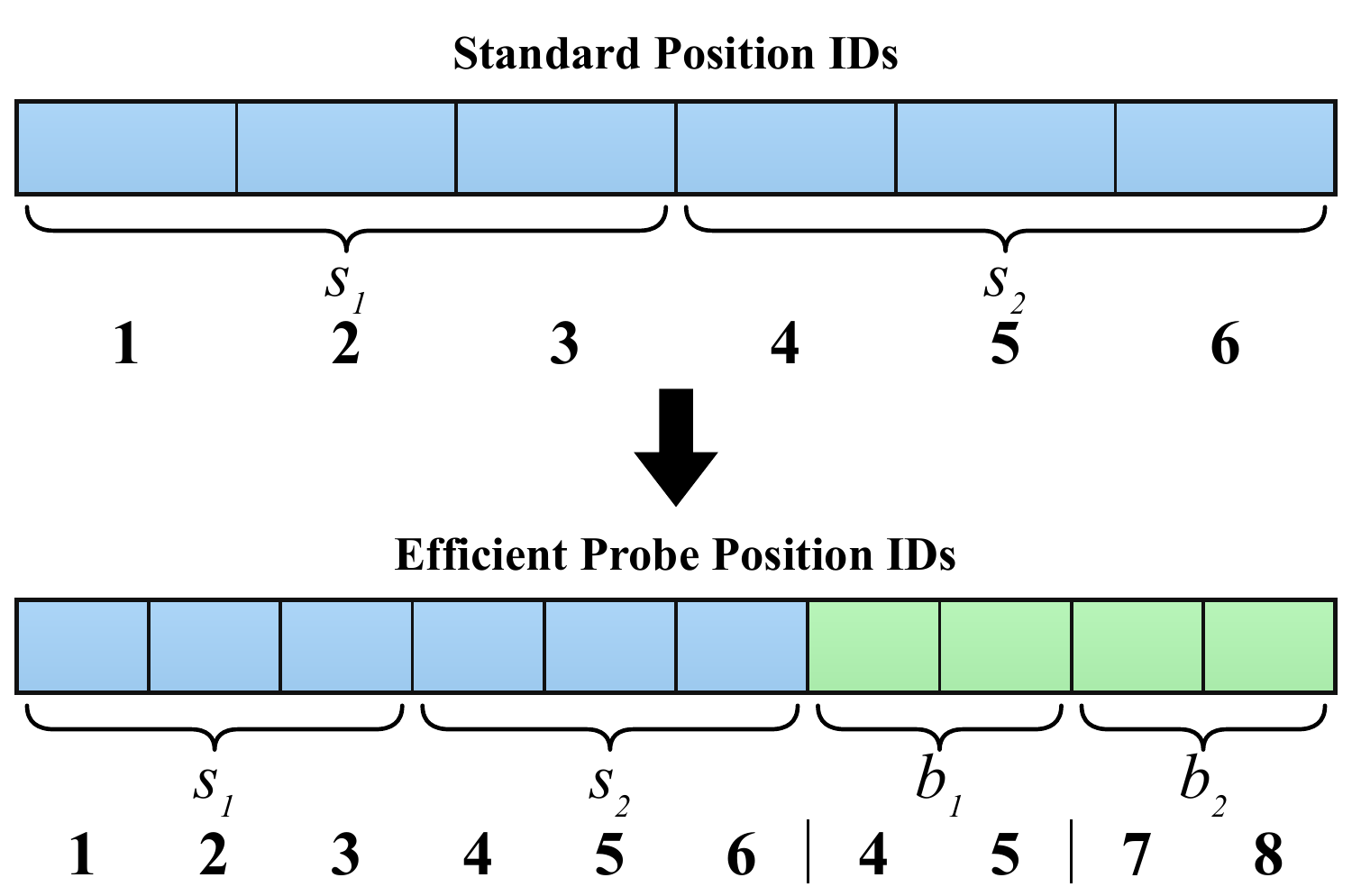}
        \\[-1pt]
        {\small (a) Efficient Probe Attention Mask}
        &
        {\small (b) Probe-Aligned Position IDs}
    \end{tabular}
    \caption{
    \textbf{Efficient answer-probe computation.}
    (a) Packed probes attend only to their reasoning prefixes.
    (b) Probe position IDs align with the corresponding reasoning steps.
    }
        \label{fig:efficient_probe}
\end{figure}

\subsection{Efficient Answer-Probe Computation}
\label{sec:efficient_probe}

Naively evaluating the answer probe at every reasoning state would require repeatedly forwarding different prefixes. Inspired by the packed probing strategy of IGPO~\citep{wang2025information}, we compute all student-probe scores within a single forward pass. For a sampled trajectory $\tau=(s_1,\ldots,s_K)$, we pack the required probe blocks after the response as $\widetilde{\tau}=(s_1,\ldots,s_K,b_0,\ldots,b_{K-1})$. Here, $b_0=(c_{\mathrm{probe}},a^\ast)$ probes the initial state before reasoning, while $b_k$ denotes the probe block after step $s_k$; we omit $b_K$ because the final step typically contains the answer and provides little additional probing value. As illustrated in Figure~\ref{fig:efficient_probe}, a customized attention mask allows each $b_k$ to attend only to the prompt, $s_{\leq k}$, and its own preceding probe tokens, while its position IDs are remapped to follow the corresponding reasoning prefix rather than its packed position. This preserves the intended prefix-conditioned probe semantics despite packing all probes into a single sequence. Student-probe scores are therefore obtained directly from the student forward pass used for optimization. Probe tokens are excluded from the training loss, so probing requires no additional model forward passes and only adds a small number of processed tokens. Appendix~\ref{app:probe_runtime} provides the runtime analysis, showing that R$^2$OPL adds only $2\%$--$8\%$ training overhead over OPD across the three training settings.

\section{Experiments}

\subsection{Experimental Setup}

\paragraph{Datasets.}
We train on a mixed dataset spanning mathematical, scientific, and logical reasoning. Specifically, we combine DAPO-17K\citep{yu2026dapo} for mathematics, the MCQ subset of Nemotron-Science-v1 for science, and the training split of LogiQA2.0\citep{liu2023logiqa} for logic, resulting in approximately 32K training examples. For evaluation, we use eight benchmarks across these three domains. Mathematical reasoning is evaluated on AMC23, AIME24, AIME25, and MATH-500\citep{lightman2024let}; scientific reasoning on GPQA-Diamond\citep{rein2023gpqa} and SciBench\citep{wang2023scibench}; and logical reasoning on LogicBench\citep{parmar2024logicbench} and LogiQA2.0\citep{liu2023logiqa}. Detailed dataset information is provided in Appendix~\ref{app:dataset_details}.


\definecolor{modelgray}{RGB}{240,240,240}
\definecolor{groupgray}{RGB}{248,248,248}
\definecolor{methodblue}{RGB}{238,245,252}
\definecolor{methodtext}{RGB}{35,85,135}

\begin{table*}[t]
\centering
\caption{
\textbf{Results across teacher--student configurations.}
The best and second-best results in each column, excluding the teacher, are
\textbf{bolded} and \underline{underlined}, respectively.
}
\label{tab:main_results}

\footnotesize
\setlength{\tabcolsep}{3.8pt}
\renewcommand{\arraystretch}{1.10}

\begin{adjustbox}{max width=\textwidth}
\begin{tabular}{@{}lccccccccc@{}}
\toprule

\textbf{Method}
&
\multicolumn{4}{c}{\textbf{Math}}
&
\multicolumn{2}{c}{\textbf{Science}}
&
\multicolumn{2}{c}{\textbf{Logic}}
&
\textbf{Avg.}
\\

\cmidrule(lr){2-5}
\cmidrule(lr){6-7}
\cmidrule(lr){8-9}

&
\textbf{AMC23}
&
\textbf{AIME24}
&
\textbf{AIME25}
&
\textbf{MATH-500}
&
\textbf{GPQA-D}
&
\textbf{SciBench}
&
\textbf{LogicBench}
&
\textbf{LogiQA2.0}
&
\\

\midrule


\rowcolor{modelgray}
\multicolumn{10}{c}{
\textbf{Qwen3-4B-Instruct-2507 $\boldsymbol{\rightarrow}$ Qwen3-1.7B}
}
\\

\addlinespace[2pt]

\rowcolor{groupgray}
\multicolumn{10}{c}{
\textit{\color{black!68}Reference Models}
}
\\[-1pt]

Student
& 44.4 & 13.8 & 9.6 & 75.4
& 31.0 & 37.5
& 79.1 & 59.8
& 43.8 \\

Teacher
& 92.8 & 59.2 & 44.6 & 96.6
& 58.5 & 63.2
& 85.2 & 80.1
& 72.5 \\

\addlinespace[3pt]

\rowcolor{groupgray}
\multicolumn{10}{c}{
\textit{\color{black!68}Reward Only}
}
\\[-1pt]

GRPO
& 64.2 & 25.6 & 16.7 & 82.9
& 33.6 & 39.8
& 79.7 & 61.4
& 50.5 \\

DAPO
& 67.2 & 31.7 & 19.4 & 83.6
& 35.1 & 39.9
& 80.2 & 66.4
& 52.9 \\

GSPO
& 60.2 & 23.5 & 14.8 & 80.8
& 33.2 & 37.9
& 78.6 & 60.3
& 48.7 \\

\addlinespace[3pt]

\rowcolor{groupgray}
\multicolumn{10}{c}{
\textit{\color{black!68}Distillation Only}
}
\\[-1pt]

OPD
& 66.7 & 32.9 & 21.0 & 86.9
& 34.6 & 42.6
& 82.0 & 65.0
& 54.0 \\

ExOPD
& 68.9 & 35.2 & 24.6 & 88.1
& 34.8 & 42.9
& 83.5 & 67.8
& 55.7 \\

EOPD
& \underline{69.7} & 33.8 & 22.7 & \underline{88.4}
& 35.6 & \underline{44.2}
& \textbf{84.0} & \underline{68.1}
& \underline{55.8} \\

\addlinespace[3pt]

\rowcolor{groupgray}
\multicolumn{10}{c}{
\textit{\color{black!68}Reward--Distillation Hybrid}
}
\\[-1pt]

SRPO
& 69.2 & \underline{35.8} & \underline{25.2} & 87.1
& \underline{35.8} & 43.7
& 82.7 & 66.6
& \underline{55.8} \\

RLSD
& 65.8 & 27.3 & 18.5 & 84.8
& 35.2 & 41.8
& 80.4 & 62.3
& 52.0 \\

OPDVR
& 67.3 & 33.5 & 21.5 & 87.3
& 34.3 & 42.6
& 81.8 & 66.3
& 54.3 \\

\addlinespace[2pt]

\rowcolor{methodblue}
\textcolor{methodtext}{\textbf{R$^{2}$OPL}}
& \textbf{78.1} & \textbf{40.6} & \textbf{28.5} & \textbf{90.6}
& \textbf{38.7} & \textbf{45.2}
& \underline{83.7} & \textbf{70.1}
& \textbf{59.4} \\

\addlinespace[4pt]
\midrule
\addlinespace[1pt]


\rowcolor{modelgray}
\multicolumn{10}{c}{
\textbf{Gemma-4-26B-A4B-it $\boldsymbol{\rightarrow}$ Gemma-4-E2B-it}
}
\\

\addlinespace[2pt]

\rowcolor{groupgray}
\multicolumn{10}{c}{
\textit{\color{black!68}Reference Models}
}
\\[-1pt]

Student
& 67.5 & 36.0 & 24.0 & 81.3
& 38.5 & 48.9
& 76.6 & 54.9
& 53.5 \\

Teacher
& 99.1 & 86.3 & 75.2 & 98.8
& 72.9 & 72.3
& 88.6 & 81.3
& 84.3 \\

\addlinespace[3pt]

\rowcolor{groupgray}
\multicolumn{10}{c}{
\textit{\color{black!68}Reward Only}
}
\\[-1pt]

GRPO
& 81.4 & 41.7 & 26.5 & 90.4
& 41.3 & 49.5
& 78.1 & 57.8
& 58.3 \\

DAPO
& 83.4 & \underline{43.5} & 29.8 & 91.1
& \textbf{44.5} & \underline{52.8}
& 79.2 & 59.4
& \underline{60.5} \\

GSPO
& 82.0 & 40.8 & 27.7 & 92.3
& 41.3 & 48.8
& 77.9 & 57.7
& 58.6 \\

\addlinespace[3pt]

\rowcolor{groupgray}
\multicolumn{10}{c}{
\textit{\color{black!68}Distillation Only}
}
\\[-1pt]

OPD
& 80.0 & 41.0 & 28.5 & 90.4
& 40.1 & 49.7
& 79.0 & 58.7
& 58.4 \\

ExOPD
& 83.1 & 42.5 & 29.8 & 91.1
& 39.7 & 48.5
& 79.0 & 59.4
& 59.1 \\

EOPD
& 82.7 & 41.9 & 30.2 & \underline{92.7}
& 40.6 & 49.2
& 79.8 & \underline{60.7}
& 59.7 \\

\addlinespace[3pt]

\rowcolor{groupgray}
\multicolumn{10}{c}{
\textit{\color{black!68}Reward--Distillation Hybrid}
}
\\[-1pt]

SRPO
& \underline{83.6} & 41.7 & \underline{31.5} & 91.3
& \underline{41.8} & 51.6
& \textbf{81.4} & 59.4
& 60.3 \\

RLSD
& 81.9 & 42.9 & 25.8 & 92.6
& 41.7 & 49.8
& 78.9 & 59.8
& 59.2 \\

OPDVR
& 80.5 & 40.8 & 26.9 & 90.2
& 39.4 & 48.6
& 79.3 & 58.0
& 58.0 \\

\rowcolor{methodblue}
\textcolor{methodtext}{\textbf{R$^{2}$OPL}}
& \textbf{87.7} & \textbf{47.5} & \textbf{32.9} & \textbf{93.3}
& \textbf{44.5} & \textbf{54.1}
& \underline{81.2} & \textbf{63.8}
& \textbf{63.1} \\

\bottomrule
\end{tabular}
\end{adjustbox}

\end{table*}

\paragraph{Models and Baselines.}
We evaluate R$^2$OPL under three teacher--student configurations: Qwen3-4B-Instruct-2507 $\rightarrow$ Qwen3-1.7B, Qwen3-4B-Instruct-2507 $\rightarrow$ Qwen3-4B\citep{yang2025qwen3}, and Gemma-4-26B-A4B $\rightarrow$ Gemma-4-E2B-it\citep{gemmateam2026gemma4}. We compare against three baseline categories under a common training and evaluation protocol. \textit{Reward-only} methods include GRPO\citep{guo2025deepseek}, DAPO\citep{yu2026dapo}, and GSPO\citep{zheng2025groupsequencepolicyoptimization}; \textit{distillation-only} methods include OPD\citep{agarwal2024policy}, ExOPD\citep{yang2026learning}, and EOPD\citep{jin2026entropy}; and \textit{reward--distillation hybrids} include SRPO\citep{li2026unifying}, RLSD\citep{yang2026self}, and OPDVR\citep{lin2026policy}. Since the teachers substantially outperform the corresponding students in our setting, we replace self-distillation in SRPO and RLSD with teacher distillation to construct stronger baselines. We report Avg@16, defined as the mean accuracy over 16 independently sampled responses per problem.

\paragraph{Implementation Details.}
We implement all methods using \texttt{verl}~\citep{sheng2025hybridflow} throughout all experiments. All experiments are conducted on the same eight NVIDIA H20 GPUs throughout. Distillation-only methods are trained for 100 steps, while reward-only and reward--distillation hybrid methods are trained for 500 steps, all with a learning rate of $1\times10^{-6}$. During training, we use a maximum response length of 8,192, temperature $1.0$, and top-$p$ $1.0$; during evaluation, these are set to 16,384, $0.6$, and $0.95$, respectively. For R$^2$OPL, we set the branch weights to $\mu=10$ and $\lambda=0.1$, the probe modulation strengths to $\alpha_R=0.25$ and $\alpha_D=0.5$. Each training batch contains 64 questions with four trajectories sampled per question, yielding 256 trajectories per optimization step. Additional implementation details and hyperparameters are provided in Appendix~\ref{app:training_details}.

\subsection{Main Results}

Table~\ref{tab:main_results} summarizes the main results across teacher--student configurations. R$^2$OPL achieves the strongest overall performance under both Qwen and Gemma settings, reaching Avg@16 scores of $59.4$ and $63.1$, respectively, and outperforming the strongest baselines by $3.6$ and $2.6$ points. The gains extend across mathematical, scientific, and logical reasoning: R$^2$OPL obtains the best or tied-best result on seven of eight benchmarks in both configurations. Notably, these improvements hold against reward-only, distillation-only, and hybrid baselines rather than arising from a single comparison category. This consistency across distinct model families, baseline types, and reasoning domains indicates that the improvements are not specific to a particular student model or task type.

The additional Qwen3-4B-Instruct-2507 $\rightarrow$ Qwen3-4B results in Appendix~\ref{app:qwen4b_results} show the same trend. R$^2$OPL reaches an Avg@16 of $73.1$, exceeding the strongest baseline by $5.1$ points while achieving the best performance on all eight benchmarks. Overall, these results demonstrate the strong and consistent performance of R$^2$OPL across teacher--student configurations.

%

\begin{table*}[t]
\centering
\caption{
\textbf{Ablation study of R$^2$OPL under the Qwen3-4B-Instruct-2507 $\boldsymbol{\rightarrow}$ Qwen3-1.7B configuration.}
The best result in each column is \textbf{bolded}.
}
\label{tab:r2opl_ablation}

\small
\setlength{\tabcolsep}{4.0pt}
\renewcommand{\arraystretch}{1.12}

\begin{adjustbox}{max width=\textwidth}
\begin{tabular}{@{}lcccccccc@{\hspace{5pt}}c@{}}
\toprule

\textbf{Variant}
&
\multicolumn{4}{c}{\textbf{Math}}
&
\multicolumn{2}{c}{\textbf{Science}}
&
\multicolumn{2}{c}{\textbf{Logic}}
&
\textbf{Avg.}
\\

\cmidrule(lr){2-5}
\cmidrule(lr){6-7}
\cmidrule(lr){8-9}

&
\textbf{AMC23}
&
\textbf{AIME24}
&
\textbf{AIME25}
&
\textbf{MATH-500}
&
\textbf{GPQA-D}
&
\textbf{SciBench}
&
\textbf{LogicBench}
&
\textbf{LogiQA2.0}
&
\\

\midrule

\rowcolor{methodblue}
\textcolor{methodtext}{\textbf{Full R$^2$OPL}}
& \textbf{78.1} & \textbf{40.6} & \textbf{28.5} & \textbf{90.6}
& \textbf{38.7} & 45.2
& 83.7 & \textbf{70.1}
& \textbf{59.4} \\

w/o failed-trajectory OPD
& 63.9 & 23.8 & 16.5 & 82.9
& 31.1 & 38.7
& 79.9 & 63.4
& 50.0 \\

w/o successful-trajectory RL
& 69.2 & 35.2 & 23.5 & 87.4
& 34.9 & 42.9
& 82.8 & 65.5
& 55.2 \\

w/o difficulty scaling
& 67.8 & 32.9 & 21.3 & 84.8
& 34.1 & 39.7
& 80.6 & 66.0
& 53.4 \\

w/o answer-probe modulation
& 77.5 & 40.2 & 26.9 & 90.1
& 36.4 & \textbf{45.7}
& \textbf{84.3} & 69.5
& 58.8 \\

\bottomrule
\end{tabular}
\end{adjustbox}

\end{table*}

\begin{figure*}[t]
    \centering
    \begin{tabular}{@{}ccc@{}}
        \includegraphics[width=0.31\textwidth]{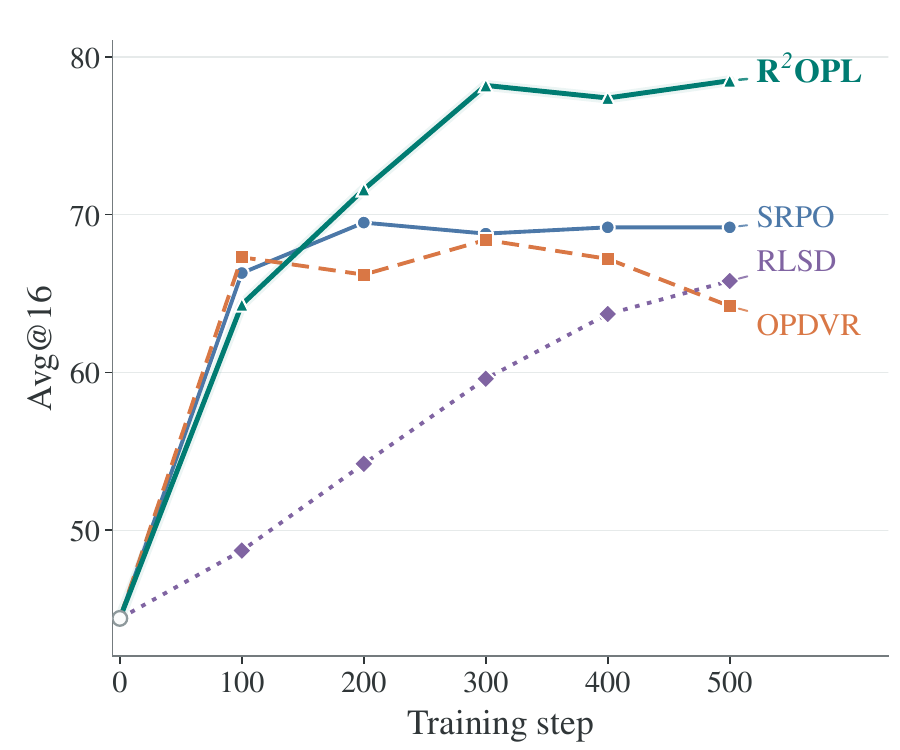}
        &
        \includegraphics[width=0.31\textwidth]{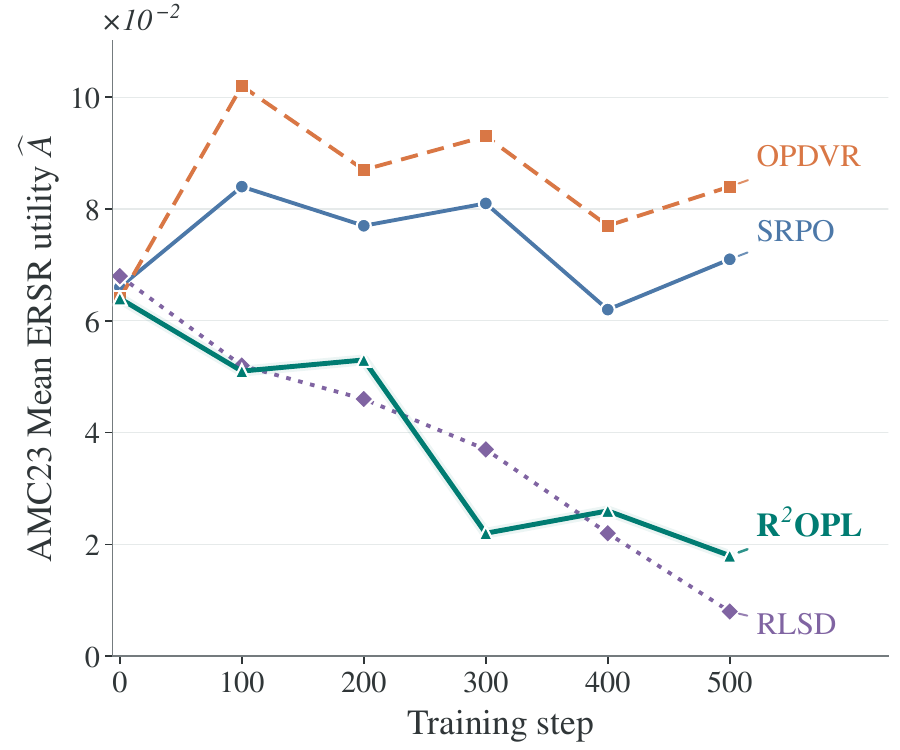}
        &
        \includegraphics[width=0.31\textwidth]{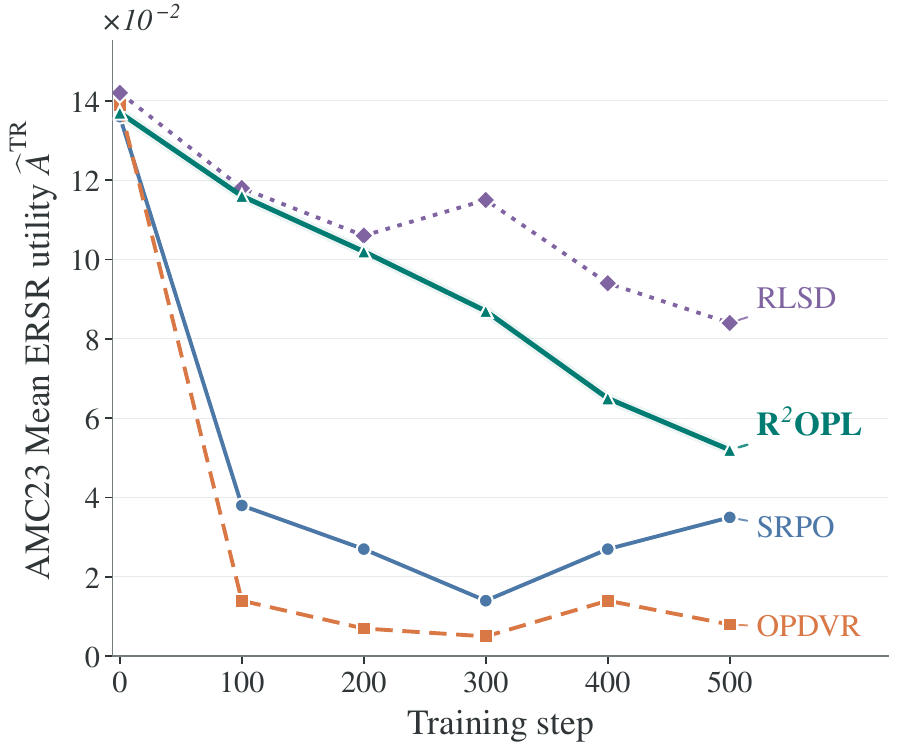}
        \\
        {\small (a) Avg@16}
        &
        {\small (b) Student-step utility $\widehat{A}$}
        &
        {\small (c) Teacher-replacement utility $\widehat{A}^{\mathrm{TR}}$}
    \end{tabular}
    \caption{
        \textbf{ERSR dynamics during training.}
        AMC23 Avg@16 and the remaining ERSR utilities of student reinforcement and teacher distillation are evaluated across training checkpoints.
    }
    \label{fig:ersr_dynamics}
\end{figure*}

\subsection{Analysis and Ablations}

\paragraph{Ablation Study.}
Table~\ref{tab:r2opl_ablation} evaluates each component of R$^2$OPL, with all variants trained under identical data, optimization, and evaluation settings. The full method achieves the best performance. Removing either failed-trajectory OPD or successful-trajectory RL causes substantial degradation, confirming their complementary roles under different trajectory outcomes. Removing difficulty scaling also reduces performance; without attenuating updates on easier prompts, overly strong positive reinforcement leads to entropy collapse, making this variant even weaker than using failed-trajectory OPD alone. Although removing answer-probe modulation yields the smallest performance drop, its best checkpoint occurs at step 500, whereas full R$^2$OPL peaks at step 250, indicating that student-probe modulation accelerates training while further improving performance.

\paragraph{ERSR Dynamics.}
Appendix~\ref{app:low_cost_ersr} shows that MC@2 over only 500 sampled steps closely approximates aggregate ERSR statistics, enabling efficient tracking during training. We use $\mathbb{E}[\widehat{A}\mid R=1]$ and $\mathbb{E}[\widehat{A}^{\mathrm{TR}}\mid R=0]$ to measure the remaining utilities of student reinforcement and teacher distillation. Their reductions during training reflect the progressive internalization of high-value actions into the policy. Figure~\ref{fig:ersr_dynamics} reveals distinct utility-consumption patterns across methods. SRPO and OPDVR rapidly improve performance while primarily exhausting teacher-distillation utility: by step 100, $\widehat{A}^{\mathrm{TR}}$ drops from $0.136$ to $0.038$ for SRPO and from $0.139$ to $0.014$ for OPDVR, while $\widehat{A}$ remains largely unconsumed. RLSD exhibits the opposite tendency, progressively reducing $\widehat{A}$ from $0.068$ to $0.008$, while retaining substantially more teacher-distillation utility. In contrast, R$^2$OPL reduces both $\widehat{A}$ from $0.064$ to $0.018$ and $\widehat{A}^{\mathrm{TR}}$ from $0.137$ to $0.052$, while achieving the highest AMC23 accuracy of $78.1$. Notably, these asymmetric utility trajectories persist even though all methods improve substantially over the initial policy. These dynamics indicate that existing hybrid methods predominantly internalize one source of expected-return utility, whereas R$^2$OPL jointly realizes both. Appendix~\ref{app:r2opl_branch_scaling} further analyzes the RL--OPD strength-control parameters $\mu$ and $\lambda$ and their effects on balancing the learning sources, measured through ERSR utility realization.

\section{Conclusion}

We introduced \textbf{Expected Reasoning-Step Return (ERSR)}, a task-grounded framework that compares and tracks student-step and teacher-replacement utility in a common expected-return space. ERSR reveals an outcome-dependent asymmetry: successful trajectories favor student reasoning, whereas failed trajectories benefit more from teacher replacement. We further showed that student answer-probe gains track student-step utility and distinguish beneficial from harmful reasoning steps. Building on these findings, we proposed \textbf{R$^2$OPL}, which combines outcome-conditioned routing, difficulty scaling, and student-probe modulation to coordinate learning from rewards and teacher signals. Experiments across reasoning domains and teacher--student configurations show strong and consistent gains, while ERSR training dynamics further reveal that R$^2$OPL jointly exploits both reward- and teacher-side utility, unlike existing hybrid methods.

\bibliography{references}
\bibliographystyle{plainnat}

\clearpage
\appendix
\section{Appendix}

\subsection{Related Work}

\paragraph{On-Policy Reinforcement Learning.}
On-policy reinforcement learning has become a major paradigm for LLM post-training, with PPO\citep{schulman2017proximal} establishing the clipped policy-gradient framework widely adopted in early RLHF systems\citep{ouyang2022training}. To reduce the computational overhead and complexity of actor--critic training, subsequent work revisits simpler critic-free policy gradients, including RLOO\citep{ahmadian2024back}, ReMax\citep{li2023remax}, and REINFORCE++\citep{hu2025reinforce++}, which improve variance reduction, training stability, and efficiency while retaining on-policy optimization. For reasoning with verifiable rewards, DeepSeekMath\citep{shao2024deepseekmath} introduced Group Relative Policy Optimization (GRPO), replacing the learned critic with relative rewards from multiple responses to the same prompt, while DeepSeek-R1\citep{guo2025deepseek} demonstrated that large-scale reinforcement learning can substantially strengthen reasoning capabilities. Recent work has refined this paradigm along several directions: DAPO\citep{yu2026dapo} improves clipping, dynamic sampling, token-level loss aggregation, and handling of overlong responses, while Lite PPO\citep{liu2025part} systematically revisits normalization, clipping, and loss aggregation; Dr.~GRPO\citep{liu2025understanding} identifies optimization biases induced by reward and length normalization, whereas Reinforce-Rej\citep{xiong2025minimalist} highlights the importance of filtering uninformative response groups. OPO\citep{hao2025policy} introduces an optimal reward baseline under exact on-policy training, while GPG\citep{chu2026gpg} directly optimizes the original policy-gradient objective without surrogate losses. Other approaches focus on stabilizing policy updates: GSPO\citep{zheng2025groupsequencepolicyoptimization} moves importance weighting and clipping to the sequence level, GMPO\citep{zhao2026geometric} uses geometric-mean aggregation to suppress outlier updates, CISPO\citep{chen2025minimax} clips importance-sampling weights rather than token updates, and SAPO\citep{gao2025soft} replaces hard clipping with smooth adaptive gating. AAPO\citep{xiong2026aapo} refines group-relative advantage estimation through an advantage-margin mechanism, while SEED-GRPO\citep{chen2025seed} modulates policy updates using semantic uncertainty. More recently, CoDaPO\citep{zhou2026easy} jointly adapts sampling and update weighting according to model confidence and problem difficulty. Despite these advances, reasoning-oriented RL methods still derive supervision primarily from final trajectory rewards, leaving credit assignment across individual reasoning steps comparatively coarse.

\paragraph{On-Policy Distillation.}
Traditional knowledge distillation transfers teacher knowledge through token- or sequence-level matching on fixed or teacher-generated data\citep{hinton2015distilling, kim2016sequence}, while LLM-oriented methods such as MiniLLM\citep{gu2024minillm} and DistiLLM\citep{ko2024distillm} improve divergence objectives and training efficiency. Generalized Knowledge Distillation (GKD)\citep{agarwal2024policy} moves distillation on-policy by training on student-generated outputs and querying the teacher on student-visited states, reducing train--inference mismatch. Recent work further studies the dynamics, efficiency, and stability of this paradigm. Rethinking OPD\citep{li2026rethinking, fu2026rethinking} analyzes teacher--student alignment and state coverage; Fast Prefix OPD\citep{zhang2026fast} focuses supervision on informative reasoning prefixes; and On-Policy Delta Distillation\citep{heo2026policy}, REOPOLD\citep{ko2026scaling}, and TrOPD\citep{xing2026trust} redesign token-level teacher signals or constrain updates under policy mismatch. A related branch replaces external teachers with privileged self-supervision: OPSD\citep{zhao2026self} conditions the same model on verified solutions, while SDPO\citep{hubotter2026reinforcement}, U-OPSD\citep{li2026policy}, and RLCSD\citep{pan2026rlcsd} extend this idea through dense self-distillation, self-consistency, or contrastive privileged signals.

Several methods combine distillation signals with outcome rewards. SRPO\citep{li2026unifying} routes successful rollouts to GRPO and failed ones to self-distillation; RLSD\citep{yang2026self} uses the privileged teacher--student gap to modulate RLVR credit; RLAD\citep{zhang2026reinforcement} selectively imitates reward-compatible teacher guidance; RG-OPD\citep{akhondzadeh2026reward} gates supervision with verifier feedback; and OPDVR\citep{lin2026policy} reshapes the implicit distillation reward by trajectory correctness. TGPO\citep{liu2026teacher} and TRAC\citep{wu2026trac} similarly integrate token-level teacher guidance with trajectory-level reward, while G-OPD\citep{yang2026learning} generalizes OPD as dense KL-constrained RL with adjustable reward scaling. Other work explores staged integration through sparse-to-dense reward transfer\citep{xu2026beyond} or sequential OPD--RL training\citep{li2026sequential}. Despite these advances, teacher supervision can remain noisy, mismatched, or overly restrictive\citep{ding2026does, yu2026mismatch, nicolicioiu2026policy, zhang2026beyond}. Existing methods therefore mainly select, filter, rescale, or combine reward- and teacher-derived signals within surrogate objectives, rather than directly comparing their utility in a common student-policy expected-return space.

\paragraph{Reasoning-Step Credit Assignment.}
Beyond trajectory-level rewards and token-level teacher supervision, recent work increasingly treats intermediate reasoning steps as the unit of learning and evaluation. Process supervision provides direct step-level correctness signals\citep{lightman2024let}, while PAVs\citep{setlur2025rewarding}, VinePPO\citep{kazemnejad2024vineppo}, and SPO\citep{guo2026segment} estimate reasoning progress or advantage from downstream success and intermediate continuations. Other methods localize influential or erroneous reasoning steps through resampling, targeted intervention, or attribution, including Self-Explore\citep{hwang2024self}, InT\citep{yang2026int}, and ACPO\citep{yin2025pinpointing}. Parallel efforts refine teacher supervision at a coarser reasoning granularity. Fast Prefix OPD\citep{zhang2026fast} and P-ALIGN\citep{liu2026long} focus distillation on informative reasoning prefixes, while TOPD\citep{jiang2026bridging} and Hindsight Self-Distillation\citep{li2026localizing} localize supervision around meaningful trajectory divergences rather than isolated token discrepancies. Relay-OPD\citep{xu2026pass} further enables localized teacher intervention when student reasoning begins to deviate. These approaches primarily improve credit assignment for student reasoning or localize teacher guidance, but evaluate them through different learning signals. ERSR instead evaluates the student step and teacher replacement from the same prefix under an identical downstream student policy, placing both in a common expected-final-reward space.

\subsection{Action Granularity for ERSR}
\label{app:token_return}

Token-level returns are formally well defined: after any generated token, one can estimate the expected final task reward under subsequent student-policy rollouts. However, formal validity does not make tokens an efficient or stable unit for reasoning credit assignment. We show that token-level estimation faces two fundamental limitations. First, when the return of a semantic reasoning operation is distributed across multiple tokens, each token carries only a small fraction of the overall effect, making its return substantially harder to resolve with finite Monte Carlo (MC) samples; meanwhile, the number of actions to be evaluated grows directly with sequence length. Second, token-level credit depends on the particular lexical realization and tokenization-induced decomposition of a reasoning operation, whereas the return of the complete semantic operation depends only on its boundary states and is invariant to its internal decomposition.

\paragraph{Monte Carlo estimation error.}
Let
\begin{equation}
    V(h)
    =
    \mathbb{E}_{\tau' \sim \pi_S(\cdot \mid h)}
    \left[R(\tau')\right]
\end{equation}
denote the expected final reward obtained by continuing from state $h$ with the student policy $\pi_S$. Since $R\in\{0,1\}$, $V(h)$ is the success probability of student-policy continuations from $h$.

Consider two post-action states $h^a$ and $h^b$, with
\begin{equation}
    p_a = V(h^a),
    \qquad
    p_b = V(h^b),
    \qquad
    \Delta = p_a-p_b.
\end{equation}
Using $N$ independent student-policy rollouts from each state, their MC estimates are
\begin{equation}
    \widehat{p}_a
    =
    \frac{1}{N}\sum_{n=1}^{N}R_n^a,
    \qquad
    \widehat{p}_b
    =
    \frac{1}{N}\sum_{n=1}^{N}R_n^b,
\end{equation}
and
\begin{equation}
    \widehat{\Delta}
    =
    \widehat{p}_a-\widehat{p}_b.
\end{equation}
The estimator is unbiased,
\begin{equation}
    \mathbb{E}[\widehat{\Delta}]
    =
    \Delta,
\end{equation}
with variance
\begin{equation}
    \operatorname{Var}(\widehat{\Delta})
    =
    \frac{
        p_a(1-p_a)+p_b(1-p_b)
    }{N}
    \leq
    \frac{1}{2N}.
    \label{eq:mc_variance}
\end{equation}
Thus, changing the action granularity does not improve the worst-case MC convergence rate: for binary final rewards, the estimation error remains of order $O(N^{-1/2})$.

What changes with granularity is the magnitude of the return difference that must be resolved. By Hoeffding's inequality,
\begin{equation}
    \Pr\left(
        |\widehat{\Delta}-\Delta|
        \geq \epsilon
    \right)
    \leq
    2\exp\left(
        -\frac{N\epsilon^2}{2}
    \right).
    \label{eq:mc_concentration}
\end{equation}
In particular, a sufficient condition for identifying the sign of a nonzero return difference with error probability at most $\alpha$ is
\begin{equation}
    N
    \geq
    \frac{2}{\Delta^2}
    \log\frac{2}{\alpha}.
    \label{eq:mc_sign_complexity}
\end{equation}
Hence, the MC budget required to distinguish beneficial from harmful actions scales quadratically with the inverse effect size,
\begin{equation}
    N
    =
    O\left(\frac{1}{\Delta^2}\right).
    \label{eq:inverse_effect_complexity}
\end{equation}

\paragraph{Statistical inefficiency at token granularity.}
Consider a reasoning step containing $m$ tokens,
\begin{equation}
    h_0
    \rightarrow
    h_1
    \rightarrow
    \cdots
    \rightarrow
    h_m,
\end{equation}
where $h_j$ denotes the state after the first $j$ tokens of the step. We define the token-level value increment and the return of the complete reasoning step as
\begin{equation}
    \delta_j = V(h_j)-V(h_{j-1}),
    \qquad
    A_{\mathrm{step}} = V(h_m)-V(h_0).
\end{equation}
These quantities satisfy the exact telescoping identity
\begin{equation}
    A_{\mathrm{step}}
    =
    \sum_{j=1}^{m}\delta_j.
    \label{eq:token_telescoping}
\end{equation}

Equation~\ref{eq:token_telescoping} does not imply that every token has a smaller effect than the complete step, since credit may concentrate on a few decisive tokens. Token-level inefficiency instead arises in the \emph{diffuse-credit} regime, where the return effect of a reasoning operation is distributed across multiple tokens. In this regime, for a substantial fraction of the tokens,
\begin{equation}
    |\delta_j|
    =
    \Theta\left(
        \frac{|A_{\mathrm{step}}|}{m}
    \right).
    \label{eq:diffuse_credit}
\end{equation}

Combining Eq.~\ref{eq:diffuse_credit} with the sign-identification bound in Eq.~\ref{eq:mc_sign_complexity}, the sufficient MC budgets for comparable confidence scale as
\begin{equation}
    N_{\mathrm{token}}
    =
    O\left(
        \frac{m^2}{A_{\mathrm{step}}^2}
        \log\frac{1}{\alpha}
    \right),
    \qquad
    N_{\mathrm{step}}
    =
    O\left(
        \frac{1}{A_{\mathrm{step}}^2}
        \log\frac{1}{\alpha}
    \right).
\end{equation}
Thus, under diffuse reasoning credit, resolving a token-level effect incurs a quadratic penalty in the number of tokens composing the step:
\begin{equation}
    \frac{N_{\mathrm{token}}}
         {N_{\mathrm{step}}}
    =
    \Theta(m^2).
    \label{eq:per_action_penalty}
\end{equation}
The difference arises not because step-level MC samples are intrinsically less noisy, but because they resolve the aggregate return effect of a complete semantic operation rather than the substantially smaller effects of its constituent lexical actions.

Token granularity incurs an additional cost when evaluating an entire response. Suppose a response contains $K$ reasoning steps with an average of $m$ tokens per step, so that $T=mK$. Step-level evaluation requires $K$ return estimates, whereas token-level evaluation considers approximately $m$ times more actions. Combining this coverage factor with the per-action penalty in Eq.~\ref{eq:per_action_penalty}, the total MC cost scales under diffuse reasoning credit as
\begin{equation}
    \frac{
        C_{\mathrm{token}}
    }{
        C_{\mathrm{step}}
    }
    =
    \Theta(m^3),
    \label{eq:total_token_penalty}
\end{equation}
up to logarithmic factors for simultaneous confidence. This reflects both the smaller return effects that must be resolved at token granularity and the substantially larger number of actions that must be evaluated.

\paragraph{Representation dependence of token-level credit.}
The limitation of token-level return is not solely statistical. Tokens are representation-dependent units whose identity and boundaries are determined by lexical realization and tokenization rather than the underlying reasoning operation. Even the same surface form may admit substantially different token decompositions. For example, \texttt{indivisible} can be decomposed as
\begin{equation}
\begin{aligned}
    &\texttt{in|div|isible},
    \qquad
    \texttt{ind|iv|isible},
    \qquad
    \texttt{indi|visible},\\
    &\texttt{in|div|is|ible},
    \qquad
    \texttt{indi|vis|ible}.
\end{aligned}
\end{equation}
These decompositions preserve the same lexical content while changing both the number and identity of token-level actions.

Representation also varies under minor surface changes that preserve essentially the same semantic function. For instance, the discourse connective \texttt{but} may correspond to distinct token forms such as
\begin{equation}
    \texttt{but},
    \qquad
    \texttt{\textvisiblespace but},
    \qquad
    \texttt{But},
    \qquad
    \texttt{\textvisiblespace But},
    \qquad
    \texttt{\textvisiblespace BUT},
\end{equation}
with further variants incorporating punctuation, such as
$\texttt{.But}$, $\texttt{,but}$, $\texttt{-but}$, and $\texttt{---but}$.
Thus, token-level actions can change with segmentation, whitespace, capitalization, or punctuation even when their underlying lexical or discourse function remains unchanged.

This creates a representation-fragmentation problem for reasoning credit. Consider a semantic reasoning operation that induces a transition from boundary state $h_{\mathrm{in}}$ to $h_{\mathrm{out}}$, with return
\begin{equation}
    A_{\mathrm{sem}}
    =
    V(h_{\mathrm{out}})
    -
    V(h_{\mathrm{in}}).
    \label{eq:semantic_return}
\end{equation}
Now consider two internal decompositions of the same transition,
\begin{equation}
\begin{aligned}
    \mathcal{P}:&\quad
    h_{\mathrm{in}}
    = h_0
    \rightarrow h_1
    \rightarrow \cdots
    \rightarrow h_m
    = h_{\mathrm{out}},\\
    \mathcal{P}':&\quad
    h_{\mathrm{in}}
    = \tilde h_0
    \rightarrow \tilde h_1
    \rightarrow \cdots
    \rightarrow \tilde h_n
    = h_{\mathrm{out}}.
\end{aligned}
\end{equation}
Their corresponding micro-action credits are
\begin{equation}
    \delta_j
    =
    V(h_j)-V(h_{j-1}),
    \qquad
    \tilde{\delta}_\ell
    =
    V(\tilde h_\ell)-V(\tilde h_{\ell-1}).
\end{equation}
The two decompositions may contain different numbers of actions and assign different values to individual actions. Nevertheless, both telescope to the same semantic-step return:
\begin{equation}
    \sum_{j=1}^{m}\delta_j
    =
    \sum_{\ell=1}^{n}\tilde{\delta}_\ell
    =
    V(h_{\mathrm{out}})
    -
    V(h_{\mathrm{in}})
    =
    A_{\mathrm{sem}}.
    \label{eq:partition_invariance}
\end{equation}

Equation~\ref{eq:partition_invariance} exposes the key distinction: micro-action credits depend on how the reasoning operation is decomposed, whereas their aggregate return depends only on its boundary states. Changing token boundaries or lexical decomposition can therefore alter the number and distribution of token-level credits without changing the utility of the complete semantic operation. Token-level attribution is representation-dependent, while semantic-step return is invariant to internal decomposition.

Taken together, these results motivate semantic reasoning steps as the action granularity for ERSR. Although token-level returns are formally well defined, finite-sample estimation must resolve smaller effects across more actions, while the resulting credit remains sensitive to lexical realization and token decomposition. Semantic steps provide a suitable intermediate granularity, retaining finer credit assignment than trajectory-level rewards while aggregating tokens into complete reasoning operations whose return effects are easier to estimate and more stable across representations.

\newtcolorbox{promptbox}[1]{
    enhanced,
    colback=gray!3,
    colframe=gray!30,
    boxrule=0.6pt,
    arc=2pt,
    left=7pt,
    right=7pt,
    top=7pt,
    bottom=6pt,
    before skip=7pt,
    after skip=8pt,
    title={#1},
    colbacktitle=methodblue,
    coltitle=methodtext,
    fonttitle=\bfseries\small,
    attach boxed title to top left={
        xshift=7pt,
        yshift=-3pt
    },
    boxed title style={
        boxrule=0pt,
        arc=2pt,
        left=6pt,
        right=6pt,
        top=2.5pt,
        bottom=2.5pt
    }
}

\subsection{Structured Prompting and Reasoning-Step Segmentation}
\label{app:step_segmentation}

\paragraph{Structured Prompting.} Reliable ERSR evaluation requires consistent boundaries between semantic reasoning steps. A standard chain-of-thought prompt encourages step-by-step reasoning but does not explicitly mark where one reasoning operation ends and the next begins, making boundary identification less reliable. We denote this prompting format as \textbf{Standard CoT}:

\begin{promptbox}{Standard CoT}
\small\ttfamily
Solve the problem step by step. Put the final answer in
\textbackslash boxed\{\{...\}\}.
\end{promptbox}

In practice, responses generated with Standard CoT often contain continuous derivations whose semantic boundaries are difficult to determine unambiguously. We therefore introduce \textbf{Explicit-Step CoT}, which requires the model to organize its reasoning using explicit step headings:

\begin{promptbox}{Explicit-Step CoT}
\small\ttfamily
Solve the problem step by step. Organize the reasoning with headings
\#\#\# Step 1, \#\#\# Step 2, and so on. Put the final answer in
\textbackslash boxed\{\{...\}\}.
\end{promptbox}

\begin{table*}[t]
\centering
\caption{
\textbf{Effect of explicit step-structured prompting on Qwen3-1.7B.}
Both prompting formats use identical decoding settings.
}
\label{tab:step_prompt_comparison}

\small
\setlength{\tabcolsep}{4.0pt}
\renewcommand{\arraystretch}{1.10}

\begin{adjustbox}{max width=\textwidth}
\begin{tabular}{@{}lcccccccc@{\hspace{5pt}}c@{}}
\toprule

\textbf{Prompt}
&
\multicolumn{4}{c}{\textbf{Math}}
&
\multicolumn{2}{c}{\textbf{Science}}
&
\multicolumn{2}{c}{\textbf{Logic}}
&
\textbf{Avg.}
\\

\cmidrule(lr){2-5}
\cmidrule(lr){6-7}
\cmidrule(lr){8-9}

&
\textbf{AMC23}
&
\textbf{AIME24}
&
\textbf{AIME25}
&
\textbf{MATH-500}
&
\textbf{GPQA-D}
&
\textbf{SciBench}
&
\textbf{LogicBench}
&
\textbf{LogiQA2.0}
&
\\

\midrule

Standard CoT
& 43.4
& 13.3
& 8.5
& \cellcolor{gray!15}75.4
& \cellcolor{gray!15}31.4
& 37.2
& 78.8
& 59.4
& 43.4 \\

Explicit-Step CoT
& \cellcolor{gray!15}44.4
& \cellcolor{gray!15}13.8
& \cellcolor{gray!15}9.6
& \cellcolor{gray!15}75.4
& 31.0
& \cellcolor{gray!15}37.5
& \cellcolor{gray!15}79.1
& \cellcolor{gray!15}59.8
& \cellcolor{gray!15}43.8 \\

\bottomrule
\end{tabular}
\end{adjustbox}

\end{table*}

\begin{table}[t]
\centering
\caption{
Compliance with Explicit-Step CoT on 1,024 sampled trajectories.
}
\label{tab:explicit_step_compliance}

\small
\setlength{\tabcolsep}{8pt}
\renewcommand{\arraystretch}{1.10}

\begin{tabular}{lcc}
\toprule
& Qwen3-1.7B & Gemma-4-E2B-it \\
\midrule
Explicit-step compliance
& 1020/1024 (99.6\%)
& 772/1024 (75.4\%) \\
\bottomrule
\end{tabular}

\end{table}

These headings make the reasoning structure explicit without sacrificing model performance. As shown in Table~\ref{tab:step_prompt_comparison}, Explicit-Step CoT matches or slightly improves Qwen3-1.7B across the evaluated benchmarks, increasing Avg@16 from $43.4$ to $43.8$. We therefore adopt Explicit-Step CoT as the unified prompting format throughout the paper, including training and evaluation for all methods as well as ERSR evaluation.

To assess how reliably different models follow the structured prompting format, we independently sample 1,024 trajectories from Qwen3-1.7B and Gemma-4-E2B-it and measure whether each response uses the required explicit step headings. As shown in Table~\ref{tab:explicit_step_compliance}, Qwen3-1.7B follows the format in $99.6\%$ of trajectories, while Gemma-4-E2B-it achieves a lower compliance rate of $75.4\%$.

\begin{table}[t]
\centering
\caption{
Reasoning-step segmentation hierarchy.
}
\label{tab:step_segmentation_rules}

\small
\setlength{\tabcolsep}{5.5pt}
\renewcommand{\arraystretch}{1.10}

\begin{tabular}{p{0.24\linewidth} p{0.31\linewidth} p{0.37\linewidth}}
\toprule
Rule & Boundary signal & Safeguard \\
\midrule

Explicit steps
& Top-level consecutive \texttt{Step N} headings
& Used whenever a valid sequence is present \\

Terminal answer
& Independent \texttt{Final Answer} heading or terminal \verb|\boxed{...}|
& Separate an independent answer section; otherwise retain the answer in the final reasoning step \\

Semantic fallback
& Repeated top-level reasoning headings
& Requires at least two complete reasoning sections and excludes non-reasoning headings \\

Conservative fallback
& No reliable structural signal
& Keep the remaining response as a single step \\

\bottomrule
\end{tabular}

\end{table}

\paragraph{Reasoning-Step Segmentation.}
As shown in Table~\ref{tab:explicit_step_compliance}, explicit \texttt{Step N} headings are nearly universal for Qwen3-1.7B but appear in only $75.4\%$ of Gemma-4-E2B-it trajectories. A segmentation procedure based solely on explicit step headings would therefore work well for Qwen but leave a substantial fraction of Gemma responses unsplit. To accommodate this model-dependent formatting behavior, we use a conservative hierarchical procedure. When a response contains a valid top-level sequence of \texttt{Step 1}, \texttt{Step 2}, $\ldots$, these headings directly define the reasoning steps. An independent \texttt{Final Answer} section is treated as a terminal segment, whereas a terminal \verb|\boxed{...}| expression contained within the final reasoning section remains part of that step. If no valid step sequence is present, we apply a semantic-heading fallback, primarily for Gemma outputs, using repeated top-level headings that denote reasoning stages such as \textit{Analysis}, \textit{Evaluation}, \textit{Inference}, \textit{Calculation}, or \textit{Conclusion}. Responses without reliable structural cues are retained as a single step to avoid speculative segmentation. Table~\ref{tab:step_segmentation_rules} summarizes the hierarchy.

To minimize spurious boundaries, we do not split on ordinary paragraphs, generic numbering, answer choices, formulas, or local subheadings. Candidate headings inside protected regions such as code blocks, mathematical environments, or tables are also ignored. For semantic fallback, short or incomplete preambles are merged with the first reasoning section, and boundaries are retained only when the resulting sections contain sufficiently complete reasoning content.

Figure~\ref{fig:segmentation_case} presents a representative Gemma-4-E2B-it response that does not follow the prescribed \texttt{Step N} format. Instead, the model organizes its reasoning through a sequence of repeated semantic headings, including \textit{Analysis}, \textit{Evaluation}, and \textit{Conclusion}. Our fallback rule recognizes these top-level headings as reliable reasoning boundaries and segments the response accordingly, while preserving the introductory preamble and terminal answer within the resulting semantic structure.

\begin{figure}[t]
\centering

\begin{tcolorbox}[
    enhanced,
    width=\linewidth,
    colback=gray!2,
    colframe=gray!35,
    boxrule=0.7pt,
    arc=2.5pt,
    left=7pt,
    right=7pt,
    top=7pt,
    bottom=7pt,
    title={Case: Semantic-Heading Fallback},
    colbacktitle=methodblue,
    coltitle=methodtext,
    fonttitle=\bfseries\normalsize,
    attach boxed title to top left={
        xshift=7pt,
        yshift=-3pt
    },
    boxed title style={
        boxrule=0pt,
        arc=2pt,
        left=7pt,
        right=7pt,
        top=3pt,
        bottom=3pt
    }
]

\scriptsize

\textbf{Question.}
In a 10-year study of 400 fifth-graders, children praised for being smart were
more likely to give up, lack confidence, and avoid risks. The same pattern was
observed across different social stages and for both men and women.
Which statement best supports this research conclusion?

\vspace{2pt}

\begin{tabular}{@{}p{0.48\linewidth}p{0.48\linewidth}@{}}
\textbf{A.} Not all praise is positive, and effects vary depending on its content and form.
&
\textbf{B.} Praised students became more risk averse and less independent.
\\[2pt]
\textbf{C.} Praise and no-praise groups showed no academic-performance difference.
&
\textbf{D.} Praise had its greatest effect on the cleverest girls.
\end{tabular}

\vspace{6pt}
{\color{gray!55}\hrule}
\vspace{6pt}

\textbf{Generated response $\rightarrow$ four semantic segments}

\vspace{4pt}

\begin{tcolorbox}[
    colback=white,
    colframe=gray!20,
    boxrule=0.45pt,
    arc=1.5pt,
    left=5pt,
    right=5pt,
    top=4pt,
    bottom=4pt
]
\textcolor{methodtext}{\textbf{Segment A}}
\hfill
{\color{gray}\textit{initial preamble}}

\vspace{2pt}
\texttt{Here is the step-by-step analysis to determine the best supporting
statement for the research conclusion.}
\end{tcolorbox}

\vspace{3pt}

\begin{tcolorbox}[
    colback=white,
    colframe=gray!20,
    boxrule=0.45pt,
    arc=1.5pt,
    left=5pt,
    right=5pt,
    top=4pt,
    bottom=4pt
]
\textcolor{methodtext}{\textbf{Segment B}}
\hfill
{\color{gray}\textit{semantic-heading boundary}}

\vspace{2pt}
\texttt{\#\#\# Analysis of the Research Conclusion}

The study concluded that kids who were praised for being smart were more likely
to give up, lack confidence, and be afraid to take risks. The finding was
repeated across different social stages and for both men and women.

The key takeaway is that praise for intelligence can have negative psychological
consequences, including giving up, reduced confidence, and risk aversion.
\end{tcolorbox}

\vspace{3pt}

\begin{tcolorbox}[
    colback=white,
    colframe=gray!20,
    boxrule=0.45pt,
    arc=1.5pt,
    left=5pt,
    right=5pt,
    top=4pt,
    bottom=4pt
]
\textcolor{methodtext}{\textbf{Segment C}}
\hfill
{\color{gray}\textit{semantic-heading boundary}}

\vspace{2pt}
\texttt{\#\#\# Evaluation of the Choices}

The response evaluates the four choices individually. Choice A emphasizes that
the effect of praise depends on its content and form; Choice B largely restates
the observed finding; Choice C concerns academic performance rather than the
reported psychological effects; and Choice D restricts the effect to a
particular subgroup. \textit{[intermediate details omitted]}
\end{tcolorbox}

\vspace{3pt}

\begin{tcolorbox}[
    colback=white,
    colframe=gray!20,
    boxrule=0.45pt,
    arc=1.5pt,
    left=5pt,
    right=5pt,
    top=4pt,
    bottom=4pt
]
\textcolor{methodtext}{\textbf{Segment D}}
\hfill
{\color{gray}\textit{semantic heading + terminal answer}}

\vspace{2pt}
\texttt{\#\#\# Conclusion}

Choice A provides the strongest explanatory context. It acknowledges that the
effect of praise depends on the specific intervention.

The correct choice is A.

\texttt{\textbackslash boxed\{A\}}
\end{tcolorbox}

\end{tcolorbox}

\caption{
\textbf{Representative semantic-heading fallback case.}
Repeated top-level headings provide semantic boundaries when no explicit \texttt{Step N} sequence is present, yielding four segments with the terminal answer retained in the final segment.
}
\label{fig:segmentation_case}

\end{figure}

\begin{table*}[t]
\centering
\caption{
Reasoning-step statistics for Qwen3-1.7B across eight evaluation benchmarks.
Statistics are computed over the number of segmented reasoning steps per trajectory.
}
\label{tab:step_statistics}

\small
\setlength{\tabcolsep}{4.0pt}
\renewcommand{\arraystretch}{1.10}

\begin{adjustbox}{max width=\textwidth}
\begin{tabular}{@{}lcccccccc@{}}
\toprule

\textbf{Statistic}
&
\multicolumn{4}{c}{\textbf{Math}}
&
\multicolumn{2}{c}{\textbf{Science}}
&
\multicolumn{2}{c}{\textbf{Logic}}
\\

\cmidrule(lr){2-5}
\cmidrule(lr){6-7}
\cmidrule(lr){8-9}

&
\textbf{AMC23}
&
\textbf{AIME24}
&
\textbf{AIME25}
&
\textbf{MATH-500}
&
\textbf{GPQA-D}
&
\textbf{SciBench}
&
\textbf{LogicBench}
&
\textbf{LogiQA2.0}
\\

\midrule

Median
& 6.0
& 8.5
& 8.0
& 5.0
& 6.0
& 5.0
& 4.5
& 4.0
\\

Mean
& 7.0
& 8.6
& 8.1
& 5.8
& 6.0
& 5.1
& 4.5
& 4.2
\\

Min.
& 1.0
& 3.0
& 4.0
& 3.0
& 3.0
& 1.0
& 2.0
& 2.0
\\

Max.
& 21.0
& 17.0
& 13.0
& 48.0
& 18.0
& 8.0
& 6.0
& 7.0
\\

\bottomrule
\end{tabular}
\end{adjustbox}

\end{table*}

\paragraph{Segmentation Statistics.}
Table~\ref{tab:step_statistics} summarizes the distribution of segmented reasoning steps produced by Qwen3-1.7B across the eight evaluation benchmarks. Mathematical tasks generally induce longer reasoning chains, with AIME24 and AIME25 reaching mean step counts of $8.60$ and $8.10$, respectively, compared with $4.50$ on LogicBench and $4.21$ on LogiQA2.0. Most benchmarks exhibit median step counts between $4$ and $8.5$, indicating that the structured prompting typically yields compact multi-step trajectories rather than excessively fragmented reasoning. The observed ranges also capture substantial variation in reasoning depth, with MATH-500 reaching up to $48$ steps while LogicBench and LogiQA2.0 remain below $8$. Overall, the statistics show that the segmentation procedure adapts naturally to the differing reasoning complexity of the evaluated domains.

\subsection{ERSR Evaluation and Additional Analyses}
\label{app:ersr_evaluation}

\subsubsection{Evaluation Pipeline}
\label{app:ersr_evaluation_pipeline}

\paragraph{Data and trajectory sampling.}
We first filter DAPO-17K to retain only English-language problems, removing Chinese examples and obtaining 14,042 questions. For each question, we sample five rollouts from Qwen3-1.7B using a maximum response length of 8,192, temperature $1.0$, and top-$p$ $1.0$, with thinking mode disabled. We use the explicit step-structured prompting format described in Appendix~\ref{app:step_segmentation} to facilitate subsequent reasoning-step segmentation. We then select 1,200 questions stratified by the within-question success rate of the five student rollouts. Specifically, we sample 200 questions from each success-rate group $\{0\%,20\%,40\%,60\%,80\%,100\%\}$, yielding 6,000 student trajectories in total. Group success rates are computed from binary answer correctness. We additionally restrict selection to questions whose five rollouts are all non-truncated, preventing truncation-induced failures from distorting the group-level success rate.

\paragraph{Reasoning-step and teacher-replacement sampling.}
We segment the 6,000 trajectories using the procedure in Appendix~\ref{app:step_segmentation} and sample approximately five reasoning steps per trajectory, resulting in 30,000 evaluated steps. We exclude the final reasoning step because it typically contains the final answer and therefore provides a less informative intervention point. For every sampled student step $s_k$, we generate a teacher replacement $s_k^T$ from Qwen3-4B-Instruct-2507 conditioned on the same prefix $h_{k-1}$. Since semantic step boundaries cannot be enforced reliably with a generation delimiter, we allow the teacher to generate up to 2,048 tokens with temperature $1.0$ and top-$p$ $1.0$, and then apply the segmentation procedure from Appendix~\ref{app:step_segmentation} to retain only the first complete teacher reasoning step. The relatively long generation budget is used to maximize the likelihood that this first semantic step is generated in full.

\paragraph{Monte Carlo return estimation.}
For each of the 30,000 sampled reasoning steps, we evaluate the three states defined by ERSR: the prefix before the student step $h_{k-1}$, the state after preserving the student step $h_k$, and the state after replacing it with the teacher step $h_k^T$. From each state, we independently sample 128 continuations using the student policy and estimate
\begin{equation}
    \widehat{V}_{k-1}
    =
    \widehat{V}_{128}(h_{k-1}),
    \qquad
    \widehat{V}_{k}
    =
    \widehat{V}_{128}(h_k),
    \qquad
    \widehat{V}^{T}_{k}
    =
    \widehat{V}_{128}(h_k^T).
\end{equation}
These estimates yield the student-step and teacher-replacement advantages,
\begin{equation}
    \widehat{A}_k
    =
    \widehat{V}_{k}-\widehat{V}_{k-1},
    \qquad
    \widehat{A}^{\mathrm{TR}}_k
    =
    \widehat{V}^{T}_{k}-\widehat{V}_{k-1}.
\end{equation}
Overall, the evaluation requires
$30{,}000\times3\times128=11.52$ million student-policy continuations, providing high-fidelity MC@128 estimates for the offline ERSR analysis.

\subsubsection{ERSR across Teachers and Problem Difficulty}
\label{app:ersr_teacher_difficulty}

\paragraph{Robustness across Teachers.}
To assess whether the observed ERSR patterns depend on a particular teacher, we additionally use Qwen3-4B to generate replacement actions for the same 30,000 sampled reasoning steps and estimate the corresponding teacher-replacement utility $\widehat{A}^{\mathrm{TR}}$ with MC@128 under the identical evaluation protocol. Table~\ref{tab:ersr_teacher_robustness} shows that both teachers exhibit the same outcome-dependent pattern: teacher replacement is more beneficial on failed than on successful trajectories. For Qwen3-4B-Instruct-2507, the mean $\widehat{A}^{\mathrm{TR}}$ increases from $+0.007$ on successful trajectories to $+0.0638$ on failed ones; Qwen3-4B shows the same direction, increasing from $+0.0146$ to $+0.0218$. This consistency indicates that the outcome-dependent ERSR asymmetry is robust to teacher choice. Moreover, Qwen3-4B-Instruct-2507 substantially outperforms Qwen3-4B as a teacher in Table~\ref{tab:qwen4b_main_results_small} and yields substantially larger replacement utility on failed trajectories. Although based on a limited set of teachers, this alignment suggests that ERSR may also reflect differences in the learning benefit available from the selected teacher.

\begin{table}[t]
\centering
\caption{
\textbf{Teacher-replacement ERSR utility across teachers.}
Values report mean $\widehat{A}^{\mathrm{TR}}$ with confidence intervals under successful and failed student trajectories.
}
\label{tab:ersr_teacher_robustness}

\small
\setlength{\tabcolsep}{7pt}
\renewcommand{\arraystretch}{1.10}

\begin{tabular}{lcc}
\toprule
\textbf{Teacher}
& $\boldsymbol{R=1}$
& $\boldsymbol{R=0}$ \\
\midrule

Qwen3-4B-Instruct-2507
& $+0.007$ [$+0.000$, $+0.013$]
& $\mathbf{+0.0638}$ [$+0.054$, $+0.075$] \\

Qwen3-4B
& $+0.0146$ [$+0.009$, $+0.020$]
& $+0.0218$ [$+0.017$, $+0.027$] \\

\bottomrule
\end{tabular}

\end{table}

\begin{figure*}[t]
    \centering
    \begin{tabular}{@{}c@{\hspace{6pt}}c@{\hspace{6pt}}c@{}}
        \includegraphics[height=0.29\textwidth,keepaspectratio]{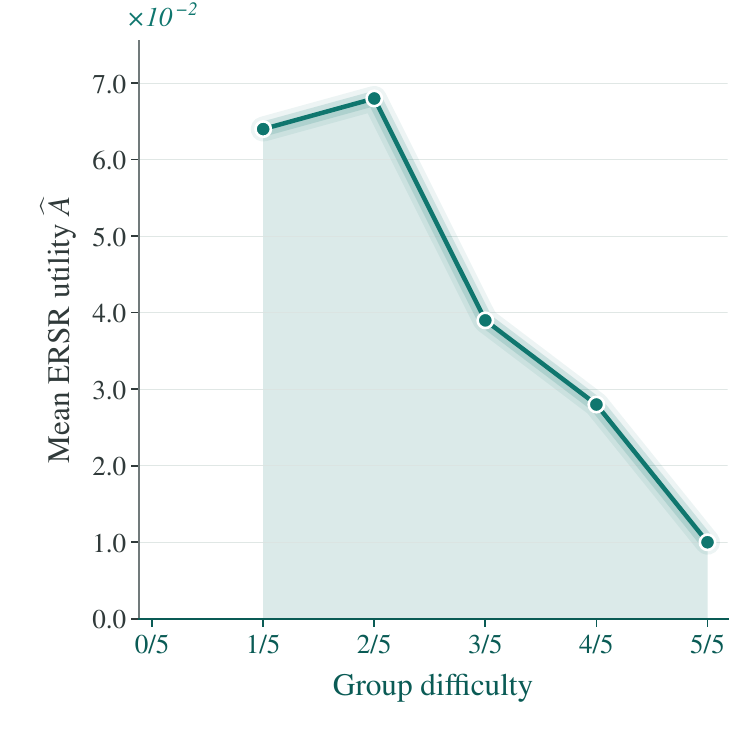}
        &
        \includegraphics[height=0.29\textwidth,keepaspectratio]{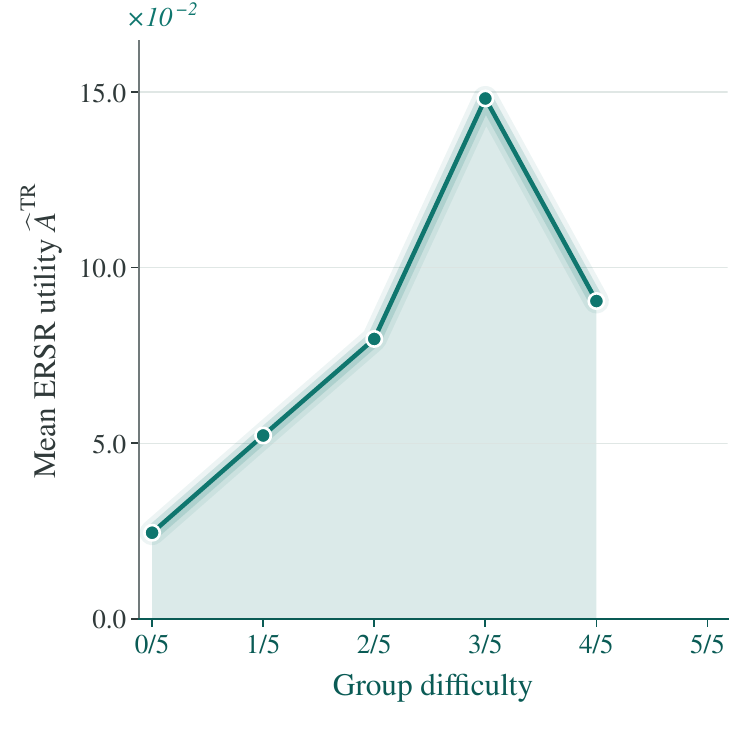}
        &
        \includegraphics[height=0.29\textwidth,keepaspectratio]{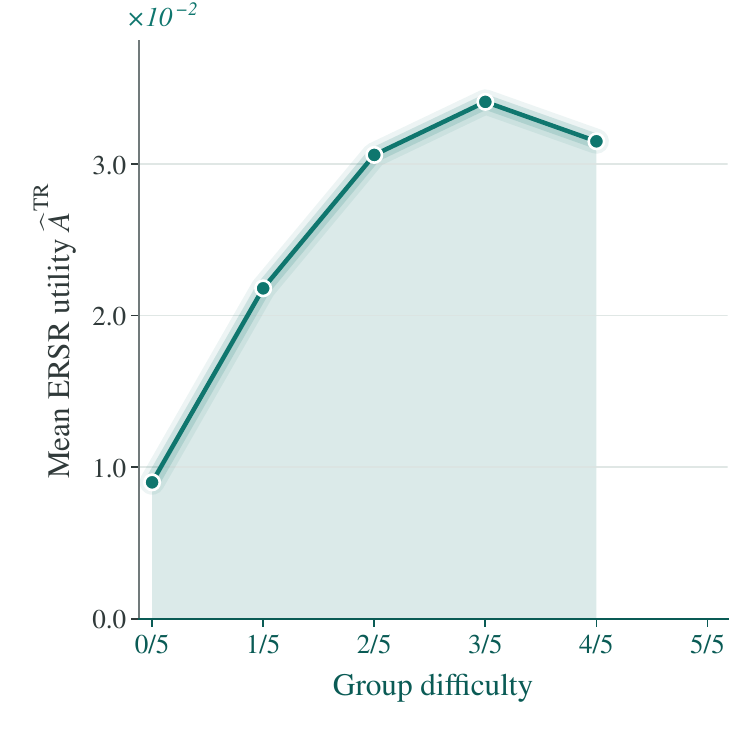}
        \\[-1pt]
        {\small (a) Student-step utility $\widehat{A}$}
        &
        {\small (b) $\widehat{A}^{\mathrm{TR}}$ with Qwen3-4B-Instruct-2507}
        &
        {\small (c) $\widehat{A}^{\mathrm{TR}}$ with Qwen3-4B}
    \end{tabular}
    \caption{
        \textbf{ERSR utility across problem difficulty.}
        Student-step utility $\widehat{A}$ is evaluated on successful trajectories, while teacher-replacement utility $\widehat{A}^{\mathrm{TR}}$ is evaluated on failed trajectories for the two teachers.
    }
    \label{fig:ersr_difficulty}
\end{figure*}

\paragraph{ERSR across problem difficulty.}
Figure~\ref{fig:ersr_difficulty} examines ERSR across groups defined by the question success rate. We compute $\widehat{A}$ only over successful trajectories ($R=1$) and $\widehat{A}^{\mathrm{TR}}$ only over failed trajectories ($R=0$); consequently, $\widehat{A}$ is undefined for the $0/5$ group, while $\widehat{A}^{\mathrm{TR}}$ is undefined for the $5/5$ group. Student-step utility generally decreases as group accuracy increases, with $\widehat{A}$ falling sharply from the intermediate-difficulty groups toward nearly zero at $5/5$, indicating diminishing benefit from self-reinforcement once the student already solves a problem reliably. Teacher-replacement utility instead exhibits a non-monotonic pattern for both teachers, peaking at intermediate difficulty. Importantly, the low $\widehat{A}^{\mathrm{TR}}$ on $0/5$ problems should not be interpreted as low value of teacher intervention. On problems where all student rollouts fail, a trajectory may contain multiple erroneous decisions, so replacing a single reasoning step is often insufficient to recover the final answer. ERSR therefore measures the local utility of a single-step replacement rather than the cumulative benefit of repeatedly distilling teacher supervision across an entire trajectory.

\subsubsection{Low-Cost ERSR Estimation}
\label{app:low_cost_ersr}

High-fidelity ERSR estimation with MC@128 is suitable for offline analysis but becomes prohibitively expensive when repeated across training checkpoints. Since our goal during training is not to recover the utility of every individual reasoning step, but to estimate the aggregate expected benefit of continued optimization, a substantially cheaper approximation may be sufficient. We therefore study whether ERSR can be reliably estimated using fewer sampled reasoning steps and a much smaller Monte Carlo budget per step, enabling practical training-time tracking of the remaining utility of reinforcement and teacher distillation.

We use the aggregate MC@128 estimates over the full ERSR evaluation set as reference values and construct low-cost estimates by subsampling fewer reasoning steps and reducing the number of continuations per state. We evaluate the branch-level quantities $\mathbb{E}[\widehat{A}\mid R=1]$ and $\mathbb{E}[\widehat{A}^{\mathrm{TR}}\mid R=0]$, rather than individual step utilities. Table~\ref{tab:low_cost_ersr} reports the resulting estimation deviations for representative budgets.

\begin{table}[t]
\centering
\caption{
\textbf{Low-cost approximation of aggregate ERSR.}
Deviations are measured against the full MC@128 estimates
($\widehat{A}=+0.032$, $\widehat{A}^{\mathrm{TR}}=+0.064$).
The shaded row denotes the configuration used for training-time tracking.
}
\label{tab:low_cost_ersr}

\small
\setlength{\tabcolsep}{5.5pt}
\renewcommand{\arraystretch}{1.12}

\begin{tabular*}{\linewidth}{
    @{\extracolsep{\fill}}
    cc
    cc
    cc
    @{}
}
\toprule

\multicolumn{2}{c}{\textbf{Estimation Budget}}
&
\multicolumn{2}{c}{$\boldsymbol{\widehat{A}}$}
&
\multicolumn{2}{c}{$\boldsymbol{\widehat{A}^{\mathrm{TR}}}$}
\\

\cmidrule(lr){1-2}
\cmidrule(lr){3-4}
\cmidrule(lr){5-6}

\textbf{Steps}
&
\textbf{MC@N}
&
\textbf{Typical Dev.}
&
\textbf{95\% Bound}
&
\textbf{Typical Dev.}
&
\textbf{95\% Bound}
\\

\midrule

250
& 4
& 0.0146
& 0.0358
& 0.0226
& 0.0554
\\

\rowcolor{gray!10}
500
& 2
& 0.0133
& 0.0326
& 0.0177
& 0.0435
\\

500
& 8
& 0.0084
& 0.0207
& 0.0150
& 0.0368
\\

1000
& 8
& 0.0060
& 0.0147
& 0.0106
& 0.0260
\\

\bottomrule
\end{tabular*}

\end{table}

The aggregate estimates remain informative even under substantially reduced budgets. Notably, 500 steps with MC@2 yields smaller deviations than 250 steps with MC@4 for both utilities despite the same nominal sampling budget, suggesting that broader coverage of reasoning steps is more valuable than allocating additional continuations to fewer steps when estimating population-level ERSR. Increasing either step coverage or the per-step MC budget further reduces error, as expected.

Based on this trade-off, we use 500 sampled steps with MC@2 for training-time ERSR tracking. As each checkpoint is independently re-estimated, these low-cost estimates may vary from the full MC@128 values due to sampling error. This approximation is intended to recover the overall remaining utility of a learning branch---whether reinforcement or teacher correction still provides substantial expected benefit---rather than to accurately estimate individual-step utilities or fine-grained subgroup statistics. We retain MC@128 for the offline ERSR analyses where such fine-grained comparisons are required.

\subsection{Answer-Probe Runtime Analysis}
\label{app:probe_runtime}

The packed answer-probe design introduces additional probe tokens into the student forward pass but does not require separate model evaluations for individual reasoning states. All probe blocks are evaluated jointly with the original trajectory using the customized attention mask and position IDs described in Section~\ref{sec:efficient_probe}, while probe tokens are excluded from the optimization loss. Consequently, the additional cost mainly arises from processing the packed probe tokens rather than repeatedly recomputing each reasoning prefix.

We measure the practical training overhead under three teacher--student configurations using identical settings for OPD and R$^2$OPL. All experiments run on the same 8$\times$NVIDIA H20 setup with a maximum response length of 8,192. We report the average wall-clock time per training step, together with the time spent on rollout generation and actor updates. Evaluation, model initialization, and checkpointing are excluded from the measurements.

\begin{table}[t]
\centering
\caption{
\textbf{Training overhead of R$^2$OPL.}
All methods use identical training settings. Times are average seconds per training step and exclude evaluation, initialization, and checkpointing.
}
\label{tab:probe_runtime}

\small
\setlength{\tabcolsep}{4.2pt}
\renewcommand{\arraystretch}{1.12}

\begin{adjustbox}{max width=\linewidth}
\begin{tabular}{@{}llcccc@{}}
\toprule

\textbf{Teacher $\rightarrow$ Student}
&
\textbf{Method}
&
\multicolumn{3}{c}{\textbf{Time per Step (s)}}
&
\textbf{Overhead}
\\

\cmidrule(lr){3-5}

&
&
\textbf{Total}
&
\textbf{Generation}
&
\textbf{Actor Update}
&
\\

\midrule

\multirow{2}{*}{Gemma-4-26B-A4B $\rightarrow$ Gemma-4-E2B-it}
& OPD
& 230.7
& 74.5
& 119.4
& -- \\

& \cellcolor{methodblue}\textcolor{methodtext}{R$^2$OPL}
& \cellcolor{methodblue}235.3
& \cellcolor{methodblue}75.3
& \cellcolor{methodblue}123.3
& \cellcolor{methodblue}$+2.0\%$ \\

\midrule

\multirow{2}{*}{Qwen3-4B-Instruct-2507 $\rightarrow$ Qwen3-1.7B}
& OPD
& 102.7
& 41.4
& 43.6
& -- \\

& \cellcolor{methodblue}\textcolor{methodtext}{R$^2$OPL}
& \cellcolor{methodblue}111.0
& \cellcolor{methodblue}43.3
& \cellcolor{methodblue}49.5
& \cellcolor{methodblue}$+8.1\%$ \\

\midrule

\multirow{2}{*}{Qwen3-4B-Instruct-2507 $\rightarrow$ Qwen3-4B}
& OPD
& 147.5
& 55.2
& 68.8
& -- \\

& \cellcolor{methodblue}\textcolor{methodtext}{R$^2$OPL}
& \cellcolor{methodblue}156.3
& \cellcolor{methodblue}55.5
& \cellcolor{methodblue}78.0
& \cellcolor{methodblue}$+6.0\%$ \\

\bottomrule
\end{tabular}
\end{adjustbox}

\end{table}

As shown in Table~\ref{tab:probe_runtime}, the packed probe computation introduces only modest wall-clock overhead across all three configurations. Total training time increases by $2.0\%$ for Gemma-4-E2B-it, $8.1\%$ for Qwen3-1.7B, and $6.0\%$ for Qwen3-4B. Generation time changes only slightly, while most of the additional cost appears in the actor update where the packed probe tokens are processed. These results show that step-level answer-probe modulation can be incorporated into on-policy training without requiring separate probe forwards, with training overhead remaining below approximately $8\%$ and dropping to as little as $2\%$ in the Gemma configuration.

\subsection{Dataset and Evaluation Details}
\label{app:dataset_details}

\paragraph{Training Data Construction.}
We construct a 32K mixed training set from DAPO-17K, Nemotron-Science-v1 MCQ, and the LogiQA2.0 training split, covering mathematical, scientific, and logical reasoning.

\begin{table}[t]
\centering
\caption{
Evaluation-set composition across mathematical, scientific, and logical benchmarks.
}
\label{tab:evaluation_data_composition}

\small
\setlength{\tabcolsep}{4.8pt}
\renewcommand{\arraystretch}{1.10}

\begin{tabular}{lcccc}
\toprule
Dataset
& Domain
& Task Format
& Original Size
& Evaluated \\
\midrule

AMC23
& Math
& Open-ended short answer
& 40
& 40 \\

AIME24
& Math
& Numerical fill-in
& 30
& 30 \\

AIME25
& Math
& Numerical fill-in
& 30
& 30 \\

MATH-500
& Math
& Open-ended short answer
& 500
& 200 \\

GPQA-Diamond
& Science
& 4-choice MCQ
& 198
& 198 \\

SciBench
& Science
& Open-ended numerical
& 580
& 200 \\

LogicBench
& Logic
& 4-choice MCQ
& 500
& 200 \\

LogiQA2.0
& Logic
& 4-choice MCQ
& 1,572
& 200 \\

\midrule
\textbf{Total}
& --
& --
& 3,450
& \textbf{1,098} \\

\bottomrule
\end{tabular}

\end{table}

\paragraph{Evaluation Set Construction.}
We evaluate on eight benchmarks spanning mathematical, scientific, and logical reasoning. To reduce evaluation cost, benchmarks containing more than 200 examples are randomly subsampled to 200 examples, while smaller benchmarks are evaluated in full, yielding 1,098 evaluation problems in total. All methods are evaluated on exactly the same subsets using identical decoding settings. Table~\ref{tab:evaluation_data_composition} summarizes the original and evaluated sizes together with the corresponding task formats. For all benchmarks except SciBench, correctness is determined by exact numerical or symbolic equivalence between the extracted \verb|\boxed{...}| answer and the reference answer, including option labels for multiple-choice tasks. For SciBench, we follow its official protocol and allow a $5\%$ numerical tolerance. When a target unit is required, we append it to the question to avoid errors caused solely by unit or scale mismatches.

\subsection{Training Configurations and Hyperparameters}
\label{app:training_details}

\begin{table}[t]
\centering
\caption{
\textbf{Shared training and evaluation configurations.}
Unless otherwise specified, all methods use the same optimization,
rollout, and evaluation settings.
}
\label{tab:shared_training_config}

\small
\setlength{\tabcolsep}{6pt}
\renewcommand{\arraystretch}{1.12}

\begin{tabular}{@{}ll@{}}
\toprule
\textbf{Hyperparameter} & \textbf{Value} \\
\midrule
Training framework & \texttt{verl} \\
Numerical precision & \texttt{bfloat16} \\
Optimizer & AdamW \\
Learning rate & $1\times10^{-6}$ \\
Learning-rate schedule & Cosine \\
Warmup & None \\
Adam coefficients & $(\beta_1,\beta_2)=(0.9,0.999)$ \\
Weight decay & 0.01 \\
Gradient clipping & 1.0 \\
KL regularization & None \\
Optimization epochs per rollout batch & 1 \\
PPO mini-batch size & 64 \\
Micro-batch size per GPU & 1 \\
Questions per batch & 64 \\
Responses per question & 4 \\
Responses per optimization step & 256 \\
Training response length & 8,192 \\
Training temperature & 1.0 \\
Training top-$p$ & 1.0 \\
Rollout backend & vLLM (asynchronous) \\
Rollout precision & \texttt{bfloat16} \\
Tensor parallel size & 1 \\
Random seed & 42 \\
Evaluation response length & 16,384 \\
Evaluation temperature & 0.6 \\
Evaluation top-$p$ & 0.95 \\
Evaluation samples & 16 (Avg@16) \\
Training steps & 100 (distillation-only); 500 (reward-based and hybrid) \\
Prompt format & Explicit-Step CoT \\
\bottomrule
\end{tabular}

\end{table}

Table~\ref{tab:shared_training_config} summarizes the shared training and evaluation configuration used across all methods. We implement all experiments with \texttt{verl} using AdamW optimization in \texttt{bfloat16} precision, a learning rate of $1\times10^{-6}$ with cosine decay and no warmup, and no auxiliary KL regularization. Each optimization step samples four responses for each of 64 prompts, yielding 256 trajectories, with a PPO mini-batch size of 64 and a per-GPU micro-batch size of 1. Rollouts are generated asynchronously with vLLM using tensor parallel size 1 and a fixed random seed of 42. All experiments are conducted on eight NVIDIA H20 GPUs. Unless otherwise specified, all methods share these settings, while method-specific hyperparameters are reported separately below.

\subsection{Results with Qwen3-4B as the Student}
\label{app:qwen4b_results}

\begin{table}[t]
\centering
\caption{
\textbf{Results on Qwen3-4B-Instruct-2507 $\boldsymbol{\rightarrow}$ Qwen3-4B.}
}
\label{tab:qwen4b_main_results_small}

\footnotesize
\setlength{\tabcolsep}{4.2pt}
\renewcommand{\arraystretch}{1.10}

\begin{adjustbox}{max width=\linewidth}
\begin{tabular}{@{}lccccccccc@{}}
\toprule

\textbf{Method}
&
\multicolumn{4}{c}{\textbf{Math}}
&
\multicolumn{2}{c}{\textbf{Science}}
&
\multicolumn{2}{c}{\textbf{Logic}}
&
\textbf{Avg.}
\\

\cmidrule(lr){2-5}
\cmidrule(lr){6-7}
\cmidrule(lr){8-9}

&
\textbf{AMC23}
&
\textbf{AIME24}
&
\textbf{AIME25}
&
\textbf{MATH-500}
&
\textbf{GPQA-D}
&
\textbf{SciBench}
&
\textbf{LogicBench}
&
\textbf{LogiQA2.0}
&
\\

\midrule

\rowcolor{modelgray}
\multicolumn{10}{c}{
\textbf{Qwen3-4B-Instruct-2507 $\boldsymbol{\rightarrow}$ Qwen3-4B}
}
\\

\addlinespace[2pt]

\rowcolor{groupgray}
\multicolumn{10}{c}{
\textit{\color{black!68}Reference Models}
}
\\[-1pt]

Student
& 65.3 & 19.6 & 18.1 & 84.8
& 40.9 & 49.6
& 84.8 & 65.8
& 53.6 \\

Teacher
& 92.8 & 59.2 & 44.6 & 96.6
& 58.5 & 63.2
& 85.2 & 80.1
& 72.5 \\

\addlinespace[3pt]

\rowcolor{groupgray}
\multicolumn{10}{c}{
\textit{\color{black!68}Reward Only}
}
\\[-1pt]

GRPO
& 80.2 & 38.5 & 27.9 & 92.2
& 46.0 & 55.8
& 84.4 & 70.9
& 62.0 \\

\addlinespace[3pt]

\rowcolor{groupgray}
\multicolumn{10}{c}{
\textit{\color{black!68}Distillation Only}
}
\\[-1pt]

OPD
& 88.0 & 49.0 & 41.9 & 93.2
& 49.6 & 59.6
& 85.9 & 73.8
& 67.6 \\

\addlinespace[3pt]

\rowcolor{groupgray}
\multicolumn{10}{c}{
\textit{\color{black!68}Reward--Distillation Hybrid}
}
\\[-1pt]

OPDVR
& 88.4 & 49.6 & 42.7 & 93.9
& 50.2 & 59.4
& 86.1 & 73.7
& 68.0 \\

\addlinespace[2pt]

\rowcolor{methodblue}
\textcolor{methodtext}{\textbf{R$^{2}$OPL}}
& \textbf{93.4}
& \textbf{55.8}
& \textbf{46.0}
& \textbf{96.9}
& \textbf{59.3}
& \textbf{65.8}
& \textbf{88.3}
& \textbf{79.5}
& \textbf{73.1} \\

\bottomrule
\end{tabular}
\end{adjustbox}

\end{table}

Table~\ref{tab:qwen4b_main_results_small} further evaluates R$^2$OPL with the stronger Qwen3-4B student. R$^2$OPL achieves an Avg@16 of $73.1$, outperforming the strongest baseline, OPDVR, by $5.1$ points and improving substantially over both GRPO ($62.0$) and standard OPD ($67.6$). The gains are consistent across all eight benchmarks, with R$^2$OPL achieving the best result among the compared training methods on every task. Notably, the optimized Qwen3-4B student also surpasses its same-scale Qwen3-4B-Instruct-2507 teacher in average performance, achieving $73.1$ versus $72.5$. It exceeds the teacher on six of eight benchmarks, including AMC23, AIME25, MATH-500, GPQA-Diamond, SciBench, and LogicBench, showing that R$^2$OPL can integrate reward-based self-improvement with teacher supervision to improve beyond the performance of the teacher itself.

\subsection{\texorpdfstring{Balancing RL and OPD in R$^2$OPL}{Balancing RL and OPD in R2OPL}}
\label{app:r2opl_branch_scaling}

\begin{figure*}[t]
    \centering
    \begin{tabular}{@{}ccc@{}}
        \includegraphics[width=0.315\textwidth]{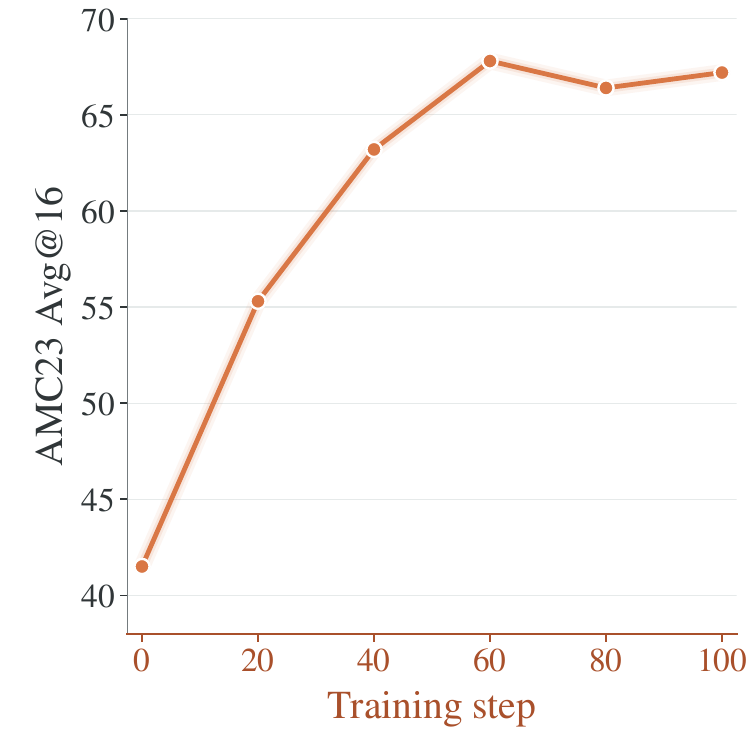}
        &
        \includegraphics[width=0.315\textwidth]{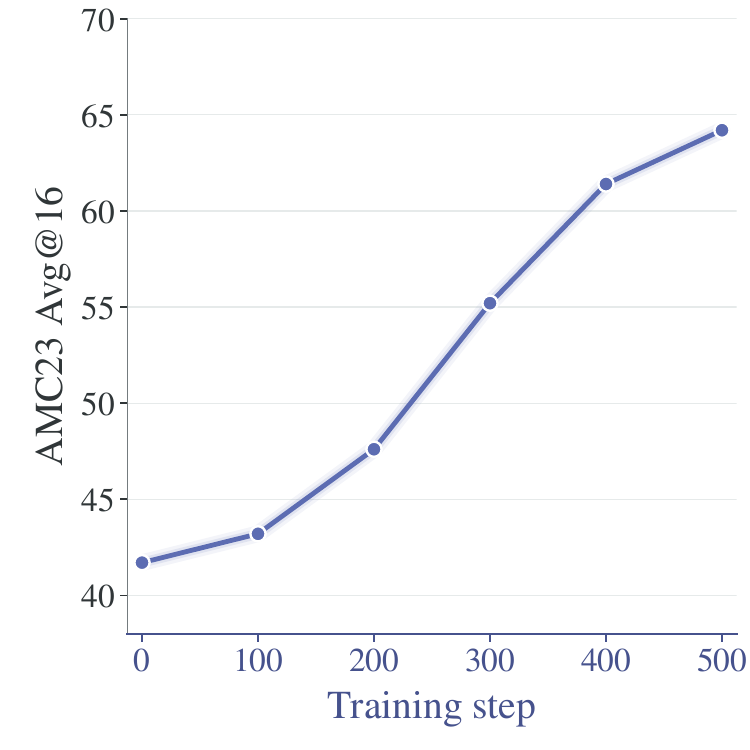}
        &
        \includegraphics[width=0.315\textwidth]{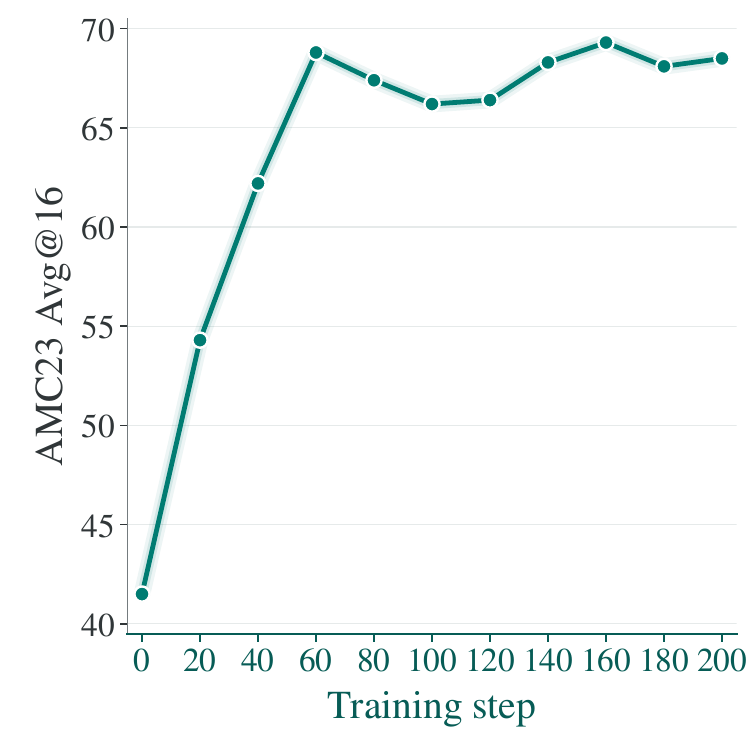}
        \\
        {\small (a) OPD, 100 steps}
        &
        {\small (b) GRPO, 500 steps}
        &
        {\small (c) R$^2$OPL ($\mu=1,\lambda=1$), 200 steps}
    \end{tabular}
    \caption{
        \textbf{AMC23 training dynamics under OPD, GRPO, and R$^2$OPL with equal branch scaling.}
        Curves report Avg@16 across training checkpoints.
    }
    \label{fig:r2opl_equal_scaling_dynamics}
\end{figure*}

\begin{figure*}[t]
    \centering
    \begin{tabular}{@{}cc@{}}
        \includegraphics[width=0.48\textwidth]{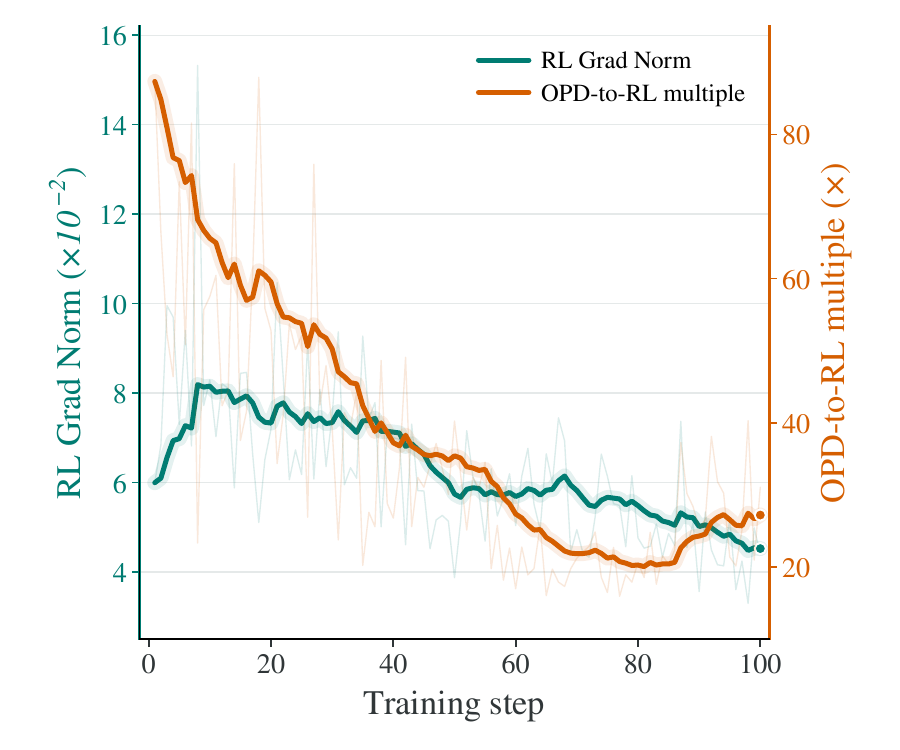}
        &
        \includegraphics[width=0.48\textwidth]{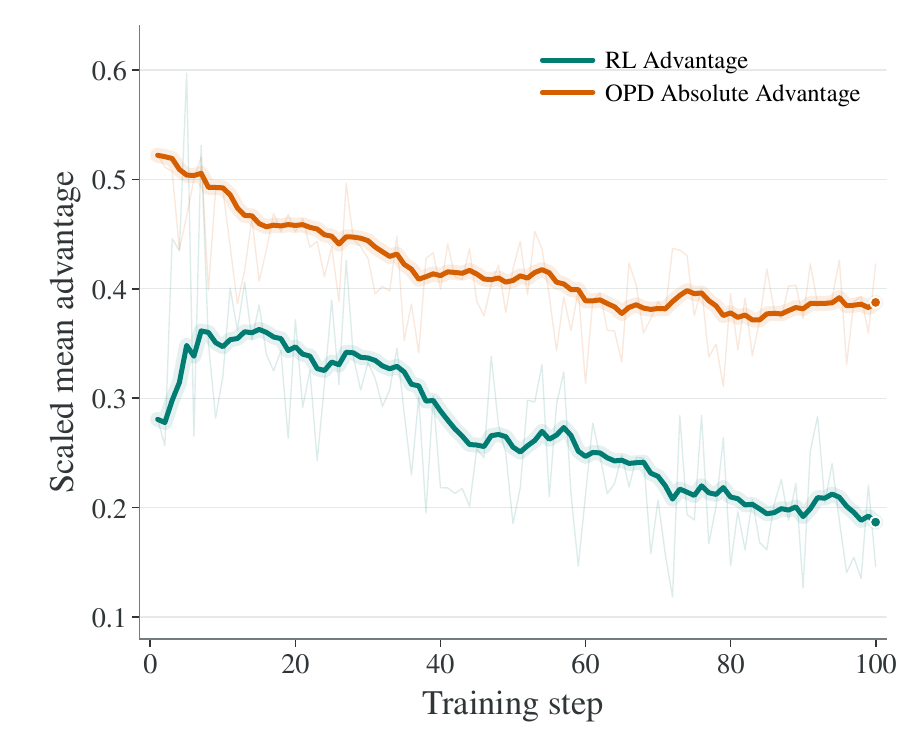}
        \\
        {\small (a) RL--OPD gradient dynamics}
        &
        {\small (b) RL--OPD advantage magnitudes}
    \end{tabular}
    \caption{
        \textbf{Branch-scale imbalance in R$^2$OPL under equal weighting ($\mu=\lambda=1$).}
        We compare the gradient norms and advantage magnitudes of the successful-trajectory RL branch and failed-trajectory OPD branch during training.
    }
    \label{fig:r2opl_equal_scaling_grad_adv}
\end{figure*}

\paragraph{Why Are $\mu$ and $\lambda$ Highly Asymmetric?}
In R$^2$OPL, $\mu$ and $\lambda$ control the relative optimization strengths of the reinforcement and distillation branches. Although a symmetric setting may appear natural, the two branches differ substantially in their learning dynamics and effective optimization influence. We therefore treat $\mu$ and $\lambda$ as balancing coefficients for the two learning processes, rather than as direct indicators of the relative importance of reward and teacher supervision.

Figure~\ref{fig:r2opl_equal_scaling_dynamics} first highlights the markedly different optimization timescales of the two learning sources. OPD improves rapidly and has largely converged by around 40 steps, whereas GRPO continues to improve through 500 steps without showing a clear convergence trend. Under the symmetric setting $\mu=\lambda=1$, R$^2$OPL closely follows the OPD-like dynamics, converging by roughly 60 steps and reaching a performance within two points of OPD. Thus, simply combining the two branches with equal coefficients largely preserves the fast convergence behavior of distillation rather than the slower continued improvement of reinforcement learning.

To understand this imbalance, Figure~\ref{fig:r2opl_equal_scaling_grad_adv} compares the branch-wise gradient norms and mean absolute advantages under $\mu=\lambda=1$. The OPD gradient norm is nearly two orders of magnitude larger than the RL gradient at the beginning of training and remains more than an order of magnitude larger throughout the first 100 steps. In contrast, the mean absolute OPD advantage is only about $1.5$--$2\times$ that of RL. The much larger disparity in gradient scale therefore cannot be explained by advantage magnitude alone, indicating that token-level OPD gradients aggregate substantially more coherently during optimization. This pronounced effective imbalance motivates substantially down-scaling the distillation branch beyond what would be suggested by the nominal advantage scales alone.

\begin{figure*}[t]
    \centering
    \begin{tabular}{@{}cc@{}}
        \includegraphics[width=0.48\textwidth]{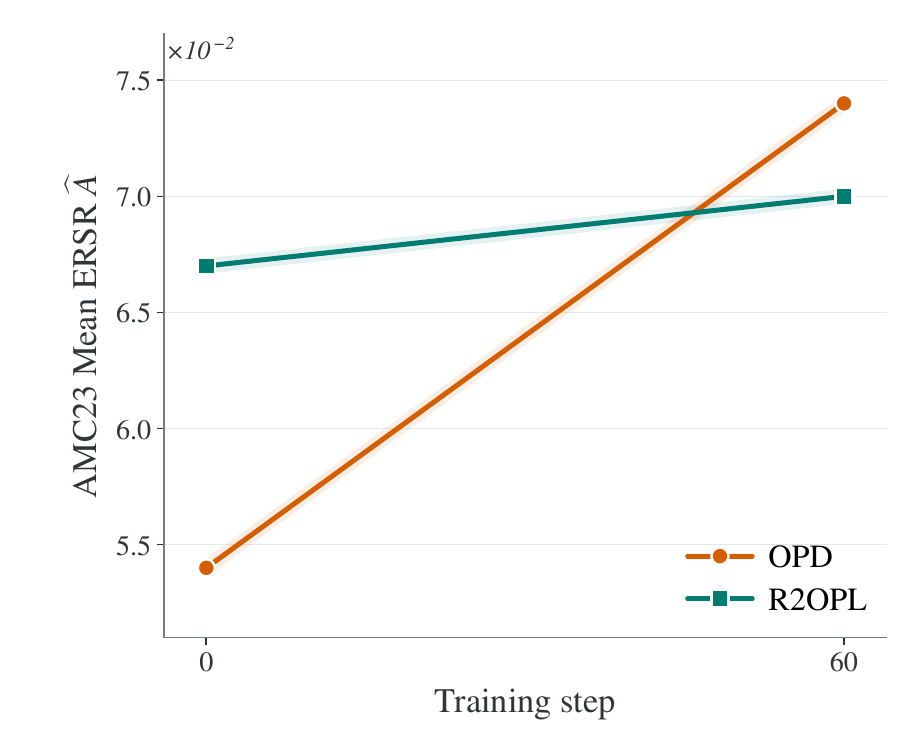}
        &
        \includegraphics[width=0.48\textwidth]{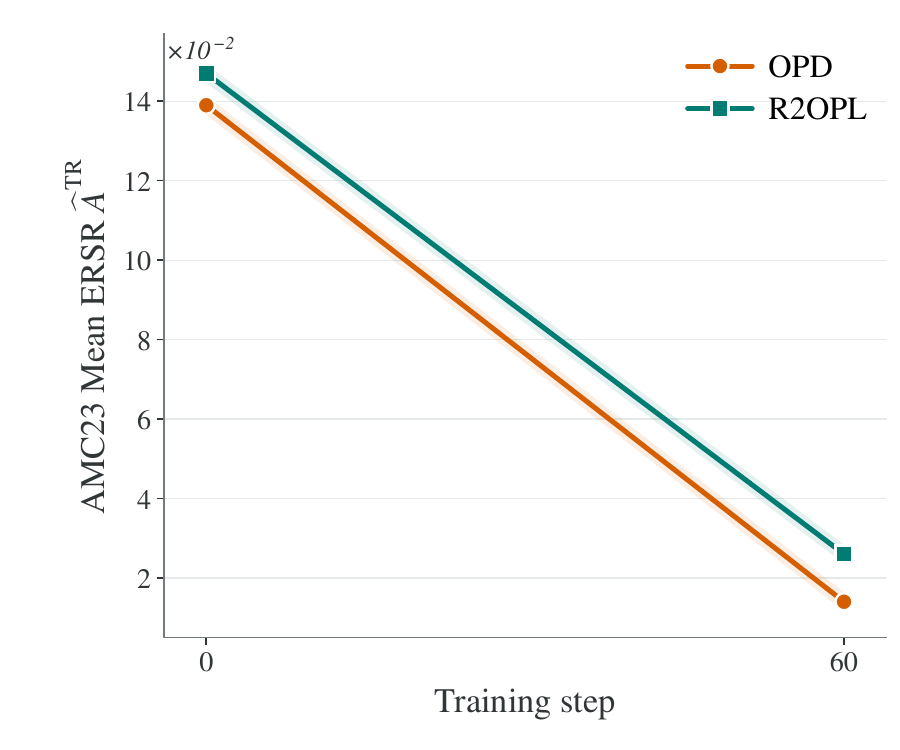}
        \\
        {\small (a) AMC23 Mean ERSR $\widehat{A}$}
        &
        {\small (b) AMC23 Mean ERSR $\widehat{A}^{\mathrm{TR}}$}
    \end{tabular}
    \caption{
        \textbf{ERSR utility dynamics under equal branch scaling ($\mu=\lambda=1$).}
        We compare the remaining student-step utility $\widehat{A}$ and teacher-replacement utility $\widehat{A}^{\mathrm{TR}}$ on AMC23 at steps 0 and 60 for OPD and R$^2$OPL.
    }
    \label{fig:r2opl_equal_scaling_ersr}
\end{figure*}

ERSR provides a direct view of which learning utility is realized under equal branch weighting. From step 0 to 60, both OPD and R$^2$OPL leave the student-step utility $\widehat{A}$ largely unchanged, while sharply reducing the teacher-replacement utility $\widehat{A}^{\mathrm{TR}}$. R$^2$OPL with $\mu=\lambda=1$ therefore exhibits essentially the same utility-consumption pattern as OPD: teacher-distillation utility is rapidly internalized, whereas reinforcement utility remains largely unexploited. Despite explicitly combining both branches, equal weighting thus leads to OPD-dominated optimization and fails to effectively internalize the student's own high-value actions.

\begin{figure*}[t]
    \centering
    \begin{tabular}{@{}ccc@{}}
        \includegraphics[width=0.315\textwidth]{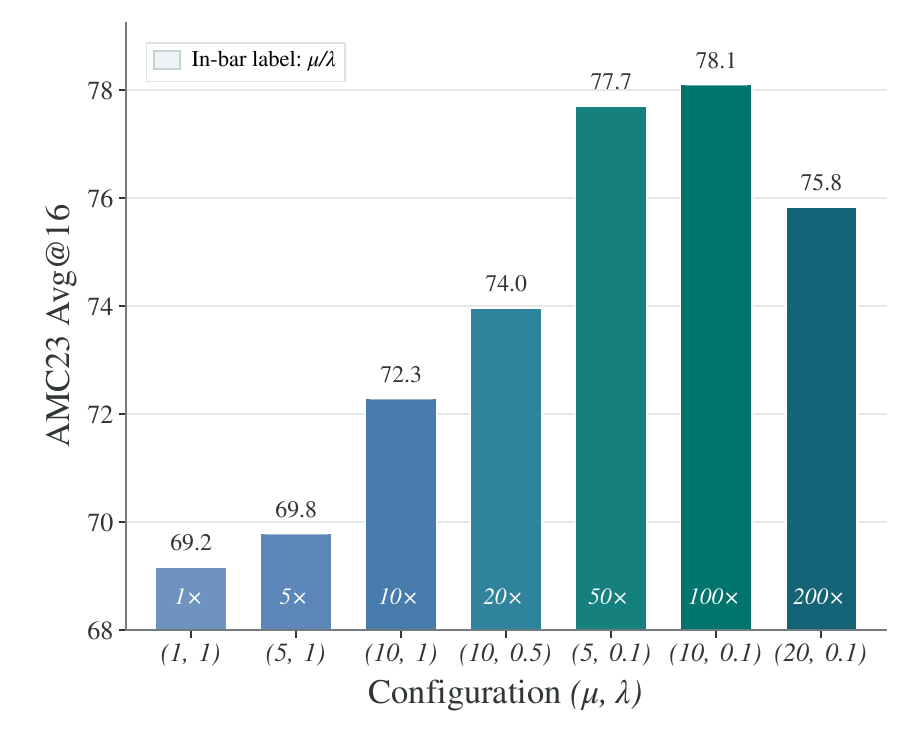}
        &
        \includegraphics[width=0.315\textwidth]{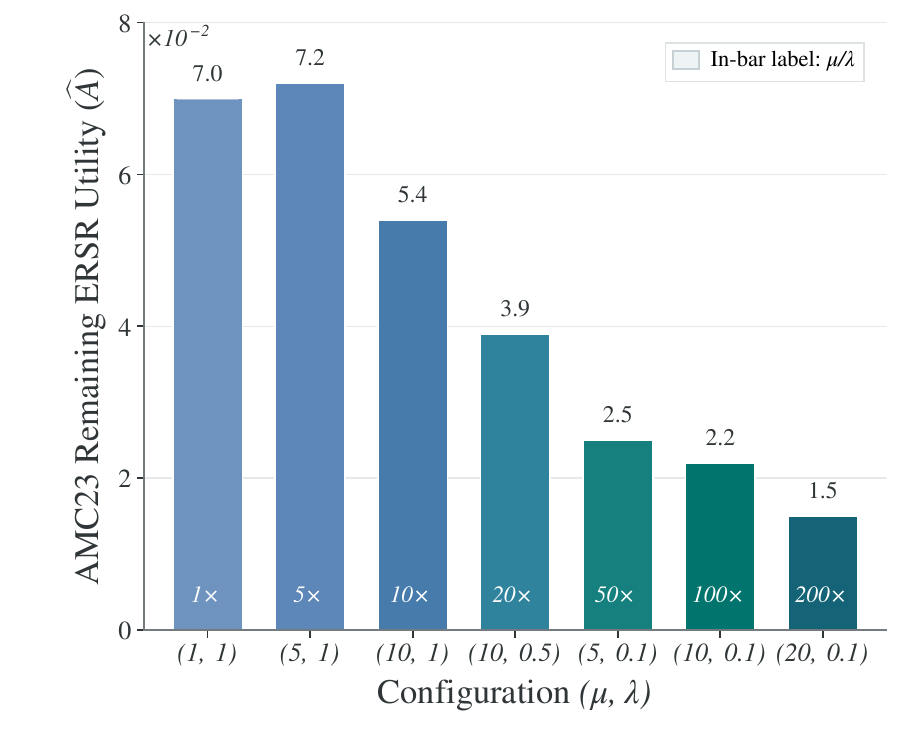}
        &
        \includegraphics[width=0.315\textwidth]{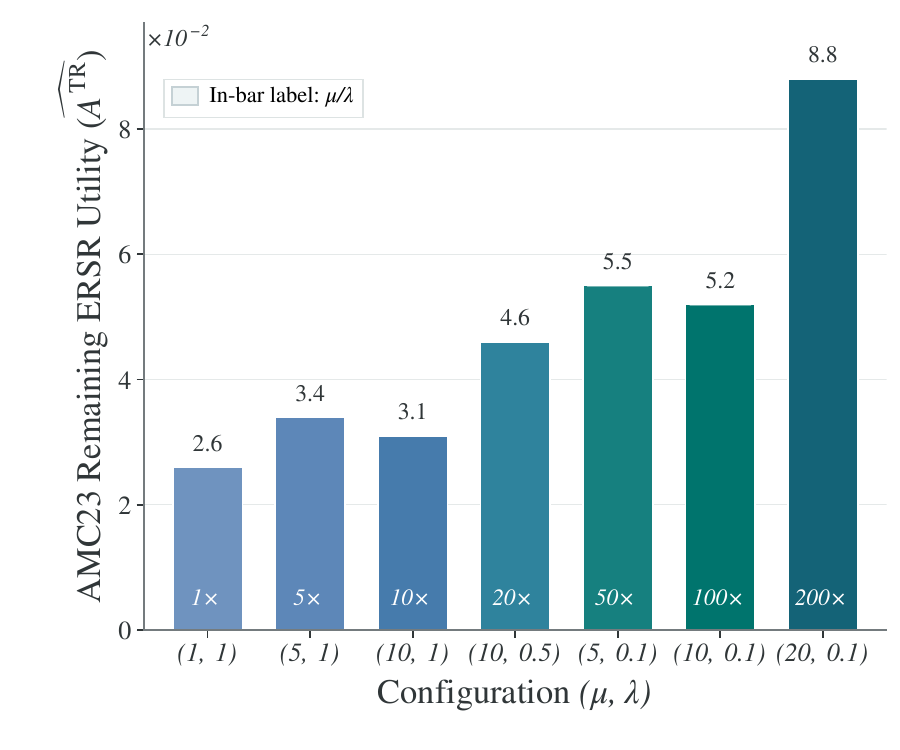}
        \\
        {\small (a) Best AMC23 Avg@16}
        &
        {\small (b) Remaining ERSR utility $\widehat{A}$}
        &
        {\small (c) Remaining ERSR utility $\widehat{A}^{\mathrm{TR}}$}
    \end{tabular}
    \caption{
        \textbf{Effect of RL--OPD branch scaling in R$^2$OPL.}
        For each $(\mu,\lambda)$ configuration, we report the best AMC23 Avg@16 over training and the remaining student-step and teacher-replacement ERSR utilities at the corresponding best checkpoint.
    }
    \label{fig:r2opl_mu_lambda_ablation}
\end{figure*}

\paragraph{Asymmetric Scaling Enables Dual Utility Realization.}
The above results show that nominally balanced RL and OPD signals can lead to strongly imbalanced optimization. We therefore vary $\mu$ and $\lambda$ over a wide range of relative branch strengths and train each configuration for 500 steps. For each setting, Figure~\ref{fig:r2opl_mu_lambda_ablation} reports the best AMC23 Avg@16 together with the remaining reinforcement utility $\widehat{A}$ and teacher-distillation utility $\widehat{A}^{\mathrm{TR}}$ at the corresponding best checkpoint. As the relative RL strength $\mu/\lambda$ increases, the remaining $\widehat{A}$ generally decreases, indicating progressively stronger realization of reinforcement utility, while $\widehat{A}^{\mathrm{TR}}$ eventually increases as teacher distillation becomes underweighted. This trade-off is reflected directly in performance: Avg@16 rises from $69.2$ at the symmetric $1\times$ setting to $78.1$ at $100\times$, where both remaining utilities are kept relatively low. Further increasing the ratio to $200\times$ reduces $\widehat{A}$ to $0.015$ but leaves substantially more teacher-distillation utility ($\widehat{A}^{\mathrm{TR}}=0.088$), accompanied by a performance drop to $75.8$. These results identify a regime in which neither branch dominates the optimization and both sources of expected-return utility are effectively internalized. We therefore use $\mu=10$ and $\lambda=0.1$ in R$^2$OPL, corresponding to a $100\times$ relative scaling between the RL and OPD branches.

This persistent OPD dominance can be understood from the directional coherence of the two learning signals. OPD provides dense token-level supervision toward a common teacher policy, causing gradients to align more consistently both within a batch and across successive batches; consequently, comparable advantage magnitudes can produce substantially larger aggregate gradients than reward-based reinforcement. This coherence also interacts with AdamW's moment accumulation. Consistently directed OPD gradients accumulate in the first moment, whereas the more variable RL gradients can partially cancel across updates while still contributing to the second moment. As a result, the effective update, which is governed by the ratio between the accumulated first and second moments, can remain strongly biased toward OPD even when its instantaneous gradient norm is smaller than that of RL. This provides an optimization-level explanation for why teacher supervision can continue to dominate joint on-policy learning after substantial down-scaling.

Unless otherwise specified, all training-time ERSR evaluations in Section~4.3 and Appendix~A.10 use 500 sampled reasoning steps with MC@2, following Appendix~A.4.3.

\end{document}